\pdfoutput=1

\documentclass[11pt]{article}

\usepackage[final]{acl}

\usepackage{times}
\usepackage{latexsym}
\usepackage{lipsum}
\usepackage{xcolor}
\usepackage{enumerate}
\usepackage{bm}

\usepackage{multirow}
\usepackage{multicol}
\usepackage{booktabs}

\usepackage[T1]{fontenc}

\usepackage[utf8]{inputenc}

\usepackage{microtype}

\usepackage{inconsolata}

\usepackage{graphicx}

\usepackage{enumitem}
\usepackage{dsfont}

\usepackage{color, colortbl}
\definecolor{Gray}{gray}{0.9}
\usepackage{multirow,adjustbox}
\usepackage{diagbox}
\newcolumntype{M}[1]{>{\centering\arraybackslash}m{#1}}

\usepackage[disable]{todonotes}
\usepackage{bm,amsmath,amssymb,fancyhdr,graphicx}
\usepackage{hyperref}
\usepackage{graphicx}
\usepackage{subcaption}

\newtheorem{definition}{Definition}[section]

\hypersetup{
  colorlinks   = true, 
  urlcolor     = purple, 
  linkcolor    = magenta, 
  citecolor   = purple 
}

\newcommand{\metric}[0]{{\tt POSIX}}

\title{\texttt{\textbf{POSIX}}: A Prompt Sensitivity Index For Large Language Models}

\author{Anwoy Chatterjee$\color{red}^{*\bm{\dagger}}$ \\
  Dept. of Electrical Engineering \\
  Indian Institute of Technology Delhi \\
  \texttt{anwoy.chatterjee@ee.iitd.ac.in} \And
  H S V N S Kowndinya Renduchintala$\color{red}^{\bm{\dagger}}$ \\
  Media and Data Science Research \\
  Adobe Inc., India \\
  \texttt{rharisrikowndinya333@gmail.com}\AND
  Sumit Bhatia \\
  Media and Data Science Research \\ 
  Adobe Inc., India \\
  \texttt{sumit.bhatia@adobe.com}\And
  Tanmoy Chakraborty \\
  Dept. of Electrical Engineering \\
  Indian Institute of Technology Delhi \\
  \texttt{tanchak@iitd.ac.in}
}

\begin{document}
\maketitle
\def\thefootnote{$\color{red}\bm{\dagger}$}\footnotetext{These two authors contributed \textbf{\color{red}equally} to this work.}
\def\thefootnote{$\color{red}*$}\footnotetext{Work done during internship at Media and Data Science Research (MDSR) Lab, Adobe Inc.}

\begin{abstract}
Despite their remarkable capabilities, Large Language Models (LLMs) are found to be surprisingly sensitive to minor variations in prompts, often generating significantly divergent outputs in response to minor variations in the prompts, such as spelling errors, alteration of wording or the prompt template. However, while assessing the quality of an LLM, the focus often tends to be solely on its performance on downstream tasks, while very little to no attention is paid to prompt sensitivity. To fill this gap, we propose \metric\ -- a novel \textit{\textbf{P}r\textbf{O}mpt \textbf{S}ensitivity \textbf{I}nde\textbf{X}} as a reliable measure of prompt sensitivity, thereby offering a more comprehensive evaluation of LLM performance. The key idea behind \metric\ is to capture the relative change in log-likelihood of a given response upon replacing the corresponding prompt with a different intent-preserving prompt. We provide thorough empirical evidence demonstrating the efficacy of \metric\ in capturing prompt sensitivity and subsequently use it to measure and thereby compare prompt sensitivity of various open-source LLMs. We find that merely increasing the parameter count or instruction tuning does not necessarily reduce prompt sensitivity whereas adding some few-shot exemplars, even just one, almost always leads to significant decrease in prompt sensitivity. We also find that alterations to prompt template lead to the highest sensitivity in the case of MCQ-type tasks, whereas paraphrasing results in the highest sensitivity in open-ended generation tasks. The code for reproducing our results is open-sourced at \href{https://github.com/kowndinya-renduchintala/POSIX}{https://github.com/kowndinya-renduchintala/POSIX}.
\end{abstract}

\section{Introduction}
\label{sec:introduction}

Large Language Models (LLMs) are pre-trained on enormous amounts of text data using the \textit{next-token-prediction} objective and they can perform a variety of NLP tasks via \textit{``prompting''}~\citep{brown2020language,kojima2022large,almazrouei2023falcon,liu2023pre,touvron2023llama}. However, LLMs have been found to be surprisingly sensitive even to the smallest of variations in prompts that do not significantly alter its meaning -- such as wording, prompt template or even minor spelling errors
-- so much so that \textit{prompt engineering}, which is a process of iteratively tuning prompts to elicit desired responses, has become a widespread practice~\cite{10.1145/3411763.3451760}.

Despite prompt sensitivity being a crucial aspect for assessing the usability of an LLM, standard evaluation benchmarks such as MMLU~\citep{mmlu} or BBH~\citep{suzgun2022challenging} focus predominantly on performance metrics like exact match,
leaving prompt sensitivity sidelined. Similarly, the model cards and blog posts announcing the most commonly used LLMs often do not contain prompt sensitivity analysis at all \cite{llama3modelcard}. However, from a user-centric perspective, models with low prompt sensitivity are generally preferred over highly prompt-sensitive ones, even if both perform similarly on standard benchmarks. This is largely because a real-world user may not always be able to formulate the "optimal" prompt everytime. Moreover, a universally applicable optimal prompt that can work \textit{across} different model architectures may not even exist. Therefore, it becomes essential to develop a dedicated procedure to systematically evaluate and quantify the sensitivity of LLMs towards \textit{intent-preserving} (or, \textit{intent-aligned}) variations in prompts.

While the exploration of the topic of quantifying prompt sensitivity is limited, there exist a few works which study prompt sensitivity and attempt to quantify it. For instance, the HELM benchmark \cite{liang2023holistic} implements various kinds of perturbations to the prompts, such as typos and misspellings, and reports the exact match score of responses obtained using the perturbed prompts. \citet{sclar2023quantifying} studied the sensitivity to variations in prompt templates. They introduced the concept of \textit{performance spread} -- defined as the difference between the highest and lowest average performances observed across a wide range of prompt templates -- and used this as a surrogate for prompt sensitivity. Likewise, ~\citet{lu2023prompts} first generated responses using variations of the same prompt and then used variation-ratio~\citep{freeman1965elementary} as the surrogate for sensitivity. 

Despite being in the right direction, the existing efforts to quantify prompt sensitivity lack nuance. Relying on variations in model accuracy as a proxy for sensitivity ignores the model's behavior in case of incorrect responses, failing to distinguish between a model that consistently generates the same incorrect response every time (\textit{low sensitivity}) and one that generates a different incorrect response with each prompt variation (\textit{high sensitivity}). Additionally, response distribution matters: a model producing a consistent response in all but one case is less sensitive than one generating two different responses - each equally frequent. The exact match score is also brittle, heavily penalizing minor wording differences in correct responses. Furthermore, models might show significant variance in the log-likelihoods of responses for different prompt variations even when the responses are identical. Finally, the existing studies are also limited to deterministic responses, like multiple-choice questions or short answers \cite{kwiatkowski-etal-2019-natural} and do not consider the sensitivity in the case of open-ended generative tasks.

With this context, in this work, we focus on the following {research question}: \textit{given a prompt along with its intent-preserving variations and the corresponding set of responses generated by a language model, how do we measure the sensitivity of the LLM on the given set of prompts such that the measure incorporates the following four key factors?
\begin{itemize}[nosep]
    \item \textbf{Response Diversity:} A higher number of unique responses generated for a given set of intent-preserving prompts should indicate higher sensitivity.
    \item \textbf{Response Distribution Entropy:} Higher entropy of the distribution of response frequencies (how often each unique response appears) should indicate higher sensitivity.
    \item \textbf{Semantic Coherence:} Lower semantic similarity among generated responses should contribute to higher sensitivity.
    \item \textbf{Variance in Confidence:} Higher variance in the log-likelihood of the same response should contribute to higher sensitivity.
\end{itemize}
}

To the best of our knowledge, ours is the first attempt towards developing a prompt sensitivity index that takes all of the above four factors into account and that works across different kinds of prompt variations and different kinds of tasks, including open-ended generation with arbitrary response lengths. 

As a first step towards addressing the research question, we identify the key property that should hold for an \textit{ideal LLM} that is not sensitive to variations in prompts, provided the underlying intent is unchanged. For this LLM, the probability of generating a certain response should remain almost the same even if minor changes are introduced in the prompt. For instance, let $x_1$ and $x_2$ be two prompts such that one is an intent-preserving variant of the other and let $y_1$ and $y_2$ be the corresponding responses generated by the LLM under consideration. Then the assumption, which turns out to be the cornerstone of our index
, is that for an LLM that is not sensitive, $\mathds{P}(y_1|x_1)$ should not change much if $x_1$ is replaced by its intent-preserving variant, $x_2$, i.e., $\mathds{P}(y_1|x_1)\approx\mathds{P}(y_1|x_2)$. Similarly the other way round --- $\mathds{P}(y_2|x_2)\approx\mathds{P}(y_2|x_1)$.

We emphasize that this notion of sensitivity does not consider the ground-truth response for quantifying sensitivity; instead, it compares the likelihood of generating various responses with different intent-preserving variants of the same prompt. In that sense, it is completely orthogonal to metrics like exact match (which are purely performance-based) and adds a new dimension altogether for evaluating LLMs.

The main contributions of our work can be summarized as follows:
\begin{itemize}[nosep]
    \item We introduce \metric\ -- a novel prompt sensitivity index that can act as a reliable measure of sensitivity of LLMs towards intent-preserving variations of prompts (Section~\ref{sec:approach}). We empirically show in Section~\ref{sec:goodness} that \metric\ incorporates all the four factors listed above.
    
    \item We use \metric\ to compute prompt sensitivity of different models and variation types, and show the effect of various aspects on the prompt sensitivity of LLMs, e.g., that increase in parameter count or instruction tuning do not necessarily decrease prompt sensitivity, and
    incorporating few-shot exemplars, even just a single exemplar, makes a huge difference in reducing prompt sensitivity (Section \ref{sec:results}).
    
    \item We reveal an interesting observation based on \metric\ computation --- variations in template bring about maximum sensitivity in the case of MCQ-type tasks while variations in wording lead to maximum effect in the case of open-ended generation tasks (Section~\ref{sec:results}).
\end{itemize}

\section{Related Work}

\noindent \textbf{Sensitivity of LLMs to Prompt Variations.} The in-context learning ability of LLMs \cite{brown2020language} makes them highly versatile, enabling them to perform a wide range of tasks through {\em prompting}, often without the need for further fine-tuning \cite{Radford2019LanguageMA, JMLR:v21:20-074, gao-etal-2021-making}. However, the robustness of in-context learning is often questioned \cite{weber2023icl}, with many studies showing that the output from LLMs is heavily dependent on aspects like the selection and ordering of in-context examples \cite{liu-etal-2022-makes, su2022selective, lu-etal-2022-fantastically, pmlr-v139-zhao21c}, choice of input labels \cite{min-etal-2022-rethinking}, or the phrasing of instruction given in the prompt \cite{gu-etal-2023-robustness, sun2024evaluating}. Apart from these aspects, LLMs are also observed to be highly sensitive to slight modifications in the structure or wordings of the prompts, even though their semantic meaning remains the same. Many prior works \cite{arora2022ask, leidinger2023language, sclar2023quantifying, voronov2024mind, mizrahi2024state} have studied this issue of sensitivity of LLMs to minor alterations in the input prompt. 

Few of these works \cite{leidinger2023language, mizrahi2024state, voronov2024mind} have also called for extending the evaluation benchmarks --
they argue that instead of evaluating on a single instance of a prompt, the benchmarks should include multiple variants for each prompt to account for the divergence in behaviour of the models to prompt variations.
While for most existing benchmarks like MMLU \cite{mmlu} or BIG-bench \cite{srivastava2022beyond, suzgun2022challenging}, the performance is reported for a single template of the prompts, the LMentry \cite{efrat-etal-2023-lmentry} benchmark uses three templates for each task in it and reports the average performance over them. Also, recently, \citet{polo2024efficient} proposed \textit{PromptEval}, a method to facilitate efficient evaluation of LLMs on any benchmark with multiple prompt templates under a limited budget. \citet{zhu2023promptbench} introduced \textit{Promptbench} for evaluating the robustness of LLMs to variations done in prompts with adversarial intent -- they observed that almost all LLMs lack robustness towards adversarial prompts. They quantified robustness using \textit{Performance Drop Rate}, which measures the relative drop in performance when perturbations are introduced into the prompt.

In this work, we also advocate for benchmarks with multiple variations of the same prompt. However, instead of relying on measures based on performance alone for measuring sensitivity or robustness, we argue the need for a comprehensive measure that can capture the prompt sensitivity of LLMs effectively.\\

\noindent \textbf{Prompt Engineering.} Due to such extensive variation in the performance of LLMs on slight modifications in input prompt, it is crucial to query LLMs with the optimal prompt to get the desired output. Prompt engineering is the practice of crafting tailored prompts for input to the LLMs to guide them towards the intended responses. Though for real-world use cases, users often perform prompt engineering manually, prior studies have also proposed ways to automate this process \cite{deng-etal-2022-rlprompt}. Another method for obtaining a better input prompt is \textit{Meta-prompting} \cite{10.1145/3411763.3451760, zhou2023large, ye2024prompt}, which aims to improve a prompt iteratively with the help of an LLM itself through further prompting. Furthermore, for tasks involving reasoning, \textit{Chain-of-Thought} prompting  \cite{wei2023chainofthought} has been observed to be very effective.

For designing a few-shot prompt, the choice of in-context examples plays a crucial role in the performance of LLMs. While existing studies \cite{liu-etal-2022-makes, min-etal-2022-rethinking} show that examples which are semantically similar to the input work the best, in some cases selecting diverse examples can be beneficial \cite{su2022selective, min-etal-2022-rethinking}. To maximize the potential of LLMs, it is therefore crucial to be aware of the best practices for prompting.

\section{\metric: Prompt Sensitivity Index}
\label{sec:approach}

Based on the notion of sensitivity introduced in Section~\ref{sec:introduction}, we will first define which intent-preserving variations we consider in this work and describe \metric\ in detail, including a brief description of the design choices involved.

\subsection{Preliminaries}
\begin{definition}
     Any two prompts $x_1$ and $x_2$ are said to be \textbf{intent-aligned} despite variations in their wording or template or inclusion of minor spelling errors, if they are designed to elicit responses from a language model based on the same underlying goal, intent or meaning.
\end{definition}

\begin{definition}
    A set of prompts $\mathbf{X}=\{x_i\}_{i=1}^{N}$ is said to be an \textbf{intent-aligned prompt set} if and only if for all $ 1\le i\neq j\le N$, $x_i\in\mathbf{X}$ and $x_j\in\mathbf{X}$ are intent-aligned.
\end{definition}

\subsection{Defining the Prompt Sensitivity Index}
\begin{definition}\label{def:index}
    Let $\mathbf{X}=\{x_i\}_{i=1}^{N}$ be an intent-aligned prompt set and $\mathbf{Y}=\{y_i\}_{i=1}^{N}$ be the set of corresponding responses generated by a language model $\mathcal{M}$, i.e., $y_i$ is the response generated by $\mathcal{M}$ when prompted using $x_i$. The sensitivity of the model $\mathcal{M}$ on $\mathbf{X}$ is defined as
    \[
        \psi_{\mathcal{M},\mathbf{X}}=\frac{1}{N(N-1)}\sum_{i=1}^{N}\sum_{j=1}^{N}\frac{1}{L_{y_j}}\left|\log\frac{\mathds{P}_{\mathcal{M}}(y_j|x_i)}{\mathds{P}_{\mathcal{M}}(y_j|x_j)}\right|
    \]
    where $N$ is the cardinality of $\mathbf{X}$, $L_{y_j}$ is the number of tokens in $y_j$, $\mathds{P}_{\mathcal{M}}(y_j|x_i)$ is the probability of the model $\mathcal{M}$ generating the response $y_j$ \textbf{given} the prompt $x_i$, and $\mathds{P}_{\mathcal{M}}(y_j|x_j)$ is the probability of the model $\mathcal{M}$ generating the response $y_j$ \textbf{given} the prompt $x_j$. 
\end{definition}

\begin{definition}
    (\textbf{\metric}) Given a language model $\mathcal{M}$ and a dataset $\mathcal{D}=\{\mathbf{X}_i\}_{i=1}^{M}$ of $M$ intent-aligned prompt sets ($X_i$'s), the prompt sensitivity index (\metric) for the language model $\mathcal{M}$ on the dataset $\mathcal{D}$ is defined as 
    \[
        \metric_{\mathcal{D}, \mathcal{M}}=\frac{1}{M} \sum_{i=1}^{M}\psi_{\mathcal{M}, \mathbf{X}_i}
    \]
\end{definition}

\begin{figure*}[htb!]
    \centering
    \begin{subfigure}[b]{3.1in}
        \centering
        \includegraphics[width=3.48in]{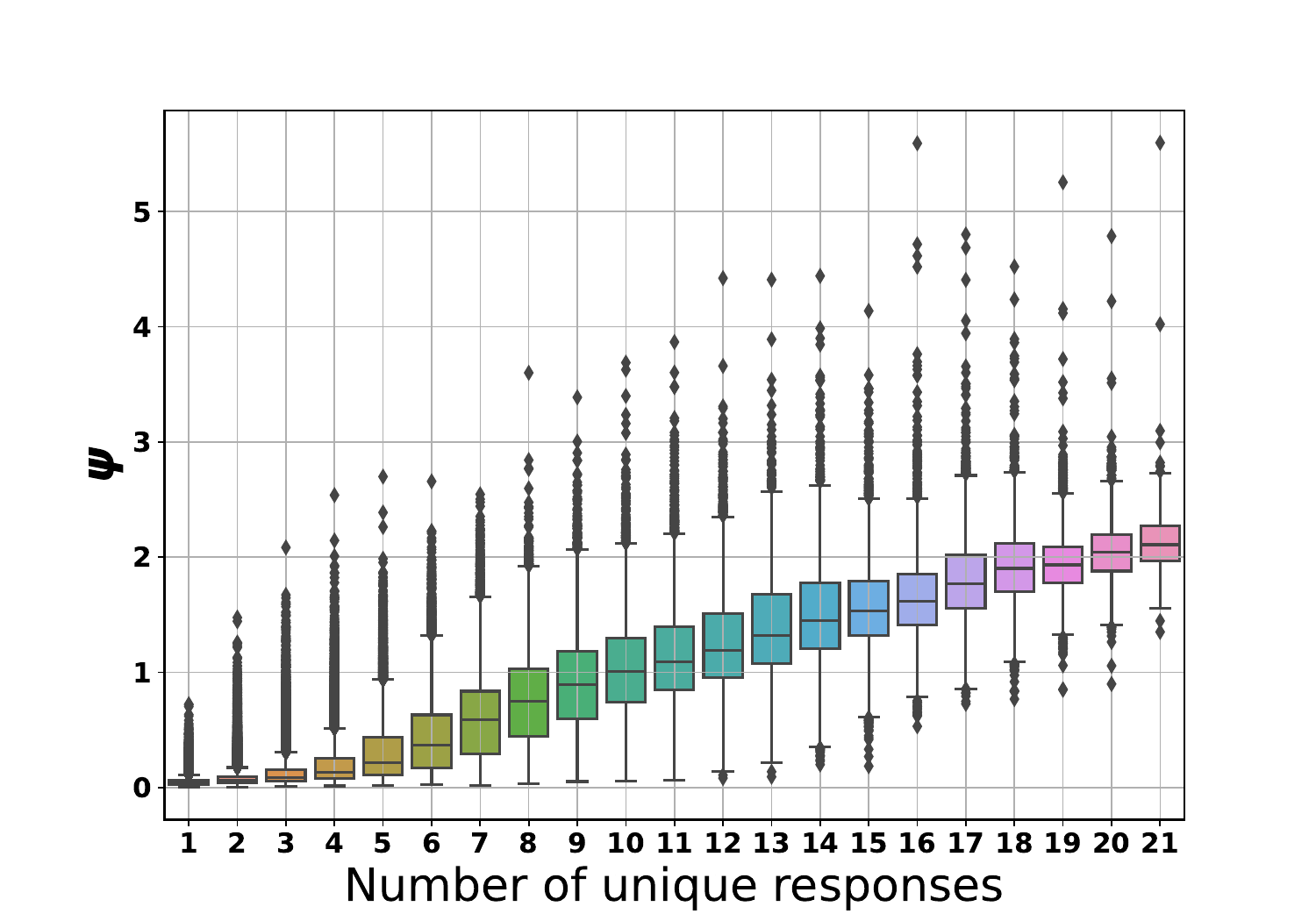}
        \caption{}
    \end{subfigure}
    \begin{subfigure}[b]{3.1in}
        \includegraphics[width=3.48in]{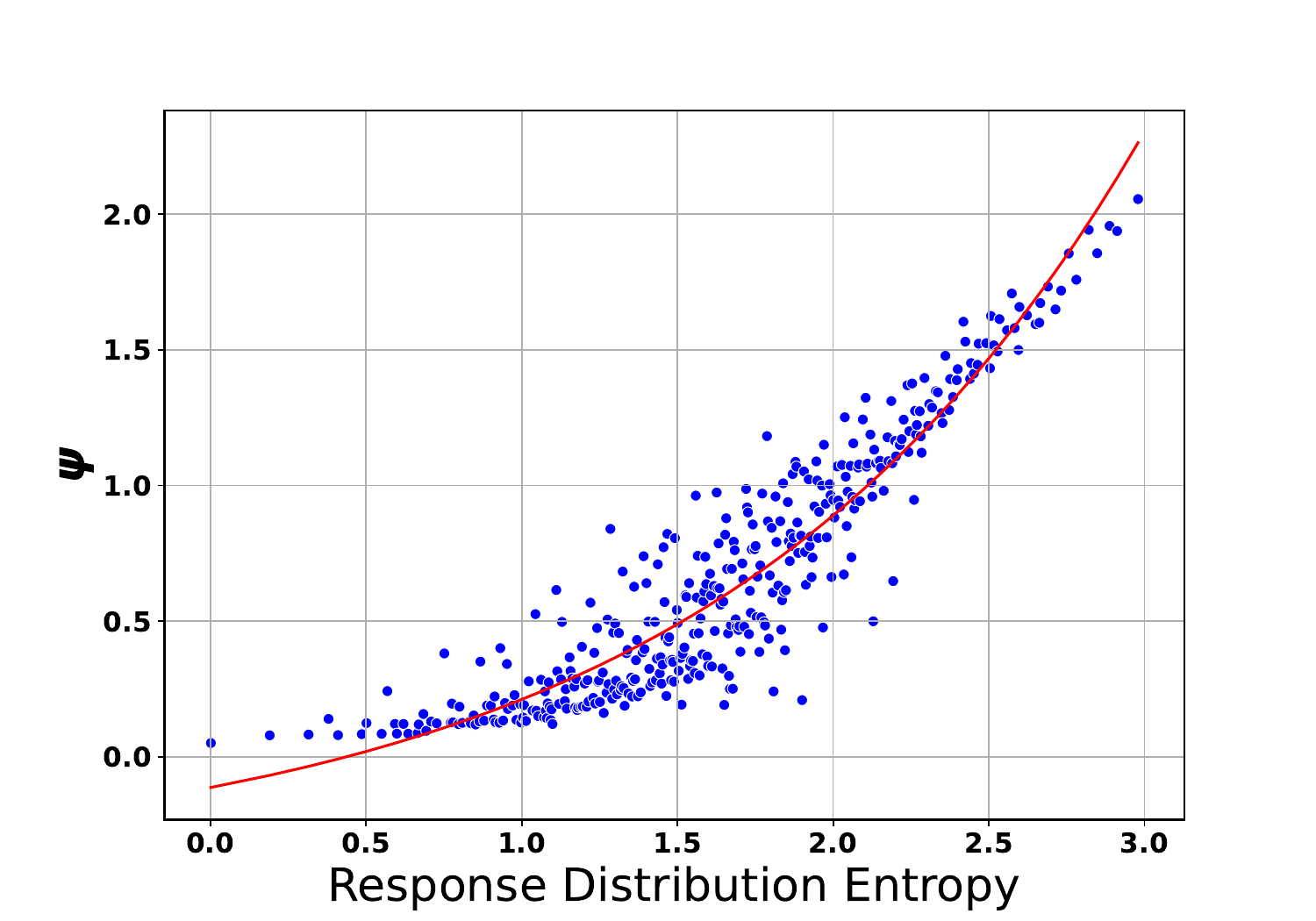}
        \caption{}
    \end{subfigure}
    \begin{subfigure}[b]{3.1in}
        \includegraphics[width=3.48in]{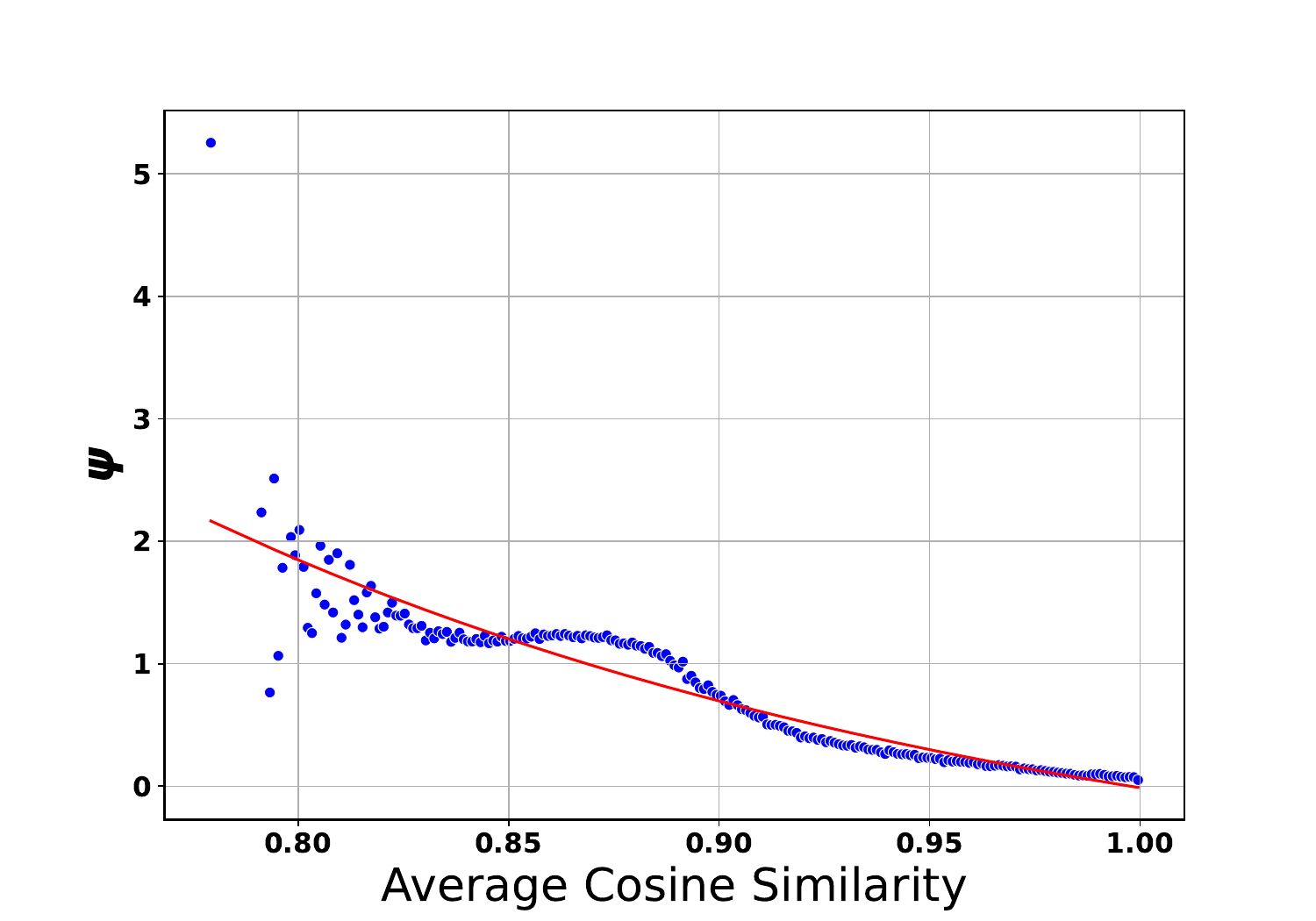}
        \caption{}
    \end{subfigure}
    \begin{subfigure}[b]{0.46\textwidth}
        \includegraphics[width=3.48in]
        {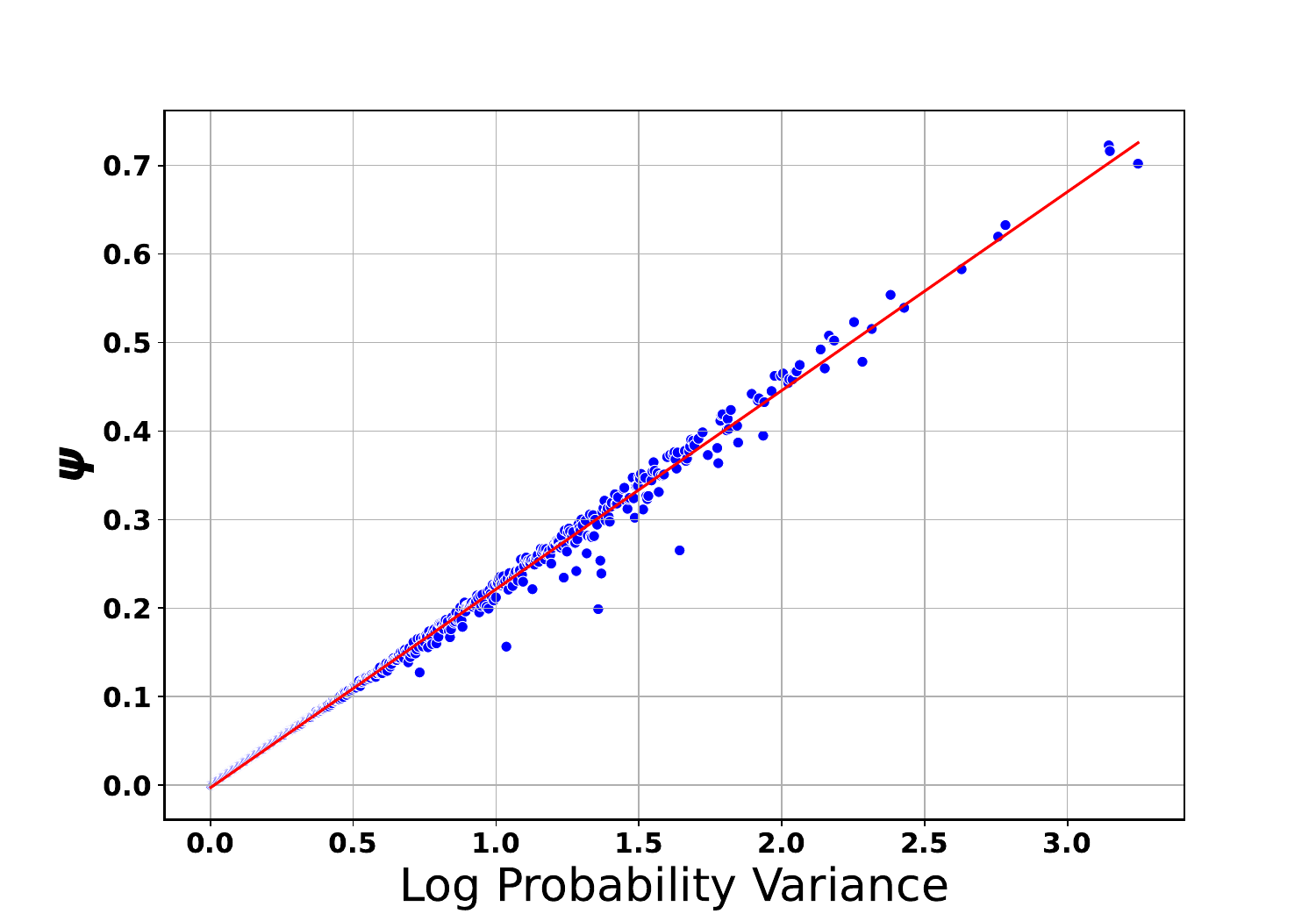}
        \caption{}
    \end{subfigure}
    \caption{Correlation plots of $\psi$ with each of the four factors described in Section~\ref{sec:what_does_posix_capture} in the case of MMLU: (a) Response Diversity; (b) Response Distribution Entropy; (c) Semantic Coherence; (d) Variance in Confidence. }
    \vspace{-3mm}
    \label{fig:efficacy_mmlu}
\end{figure*}

\subsection{What does $\psi_{\mathcal{M}, \mathbf{X}}$ Capture?}
\label{sec:what_does_posix_capture}
As mentioned briefly in Section~\ref{sec:introduction}, if we have two intent-aligned prompts $x_i$ and $x_j$ and the corresponding responses $y_i$ and $y_j$ generated by an LLM $\mathcal{M}$, we would essentially like to capture how different are $\mathds{P}_{\mathcal{M}}(y_i|x_i)$ and $\mathds{P}_{\mathcal{M}}(y_i|x_j)$ (and also similarly for $\mathds{P}_{\mathcal{M}}(y_j|x_j)$ and $\mathds{P}_{\mathcal{M}}(y_j|x_i)$). In order to make the sensitivity measure comparable across different intent-aligned prompt sets as well as across models, we need to remove the dependence on the overall scale of the model's probability distributions. Therefore, we consider the ratios $\frac{\mathds{P}_{\mathcal{M}}(y_j|x_i)}{\mathds{P}_{\mathcal{M}}(y_j|x_j)}$ which are immune to scale. And, since we only need to look at relative change in probabilities by replacing $x_i$ with $x_j$ or vice-versa, we convert the probability ratio to the logarithmic scale and also use the absolute value on top of it. Furthermore, in order to accommodate for arbitrary response lengths, we use length normalization for each term in the summation.

We now look at conceptual explanations for why $\psi_{\mathcal{M}, \mathbf{X}}$ incorporates four properties listed in Section~\ref{sec:introduction} while deferring the empirical evidence of the same to Section~\ref{sec:goodness}.
\subsubsection{\metric\ and Response Diversity}
\label{sec:approach_response_diversity}
While not explicitly evident in the expression of $\psi_{\mathcal{M}, \mathbf{X}}$, response diversity contributes indirectly to higher $\psi_{\mathcal{M}, \mathbf{X}}$. 
Say if two responses, \(y_i\) and \(y_j\), are significantly different, then their log-likelihoods (given a prompt) are likely to be significantly different i.e., the terms in summation,  $\left|\log\frac{\mathds{P}_{\mathcal{M}}(y_j|x_i)}{\mathds{P}_{\mathcal{M}}(y_j|x_j)}\right|$,  become large, thereby leading to greater $\psi_{\mathcal{M}, \mathbf{X}}$ overall. 

\subsubsection{\metric\ and Response Distribution Entropy}
\label{sec:approach_response_distribution_entropy}
By response distribution entropy, we mean the entropy of the distribution of response frequencies, i.e., how many times each unique response appears in $\mathbf{Y}$. A higher entropy 
indicates the tendency of the model to generate divergent responses more often -- consequently, the magnitude of the log-likelihood ratios in the summation of Definition \ref{def:index} will tend to be high, resulting in an uptick in the value of $\psi_{\mathcal{M}, \mathbf{X}}$ with increase in response distribution entropy.

\subsubsection{\metric\ and Semantic Coherence}
\label{sec:approach_semantic_coherence}
When the responses to intent-aligned prompts are semantically similar, i.e., the average cosine similarity between their embeddings is high, then intuitively the model is less sensitive.
This is also captured by $\psi_{\mathcal{M}, \mathbf{X}}$ because if $x_i$ and $x_j$ both generate semantically similar responses $y_i$ and $y_j$, then the probability of generating $y_j$ typically does not differ significantly for the two intent-aligned prompts $x_i$ and $x_j$ and so in such cases, the individual terms in the summation are low, leading to a lower $\psi_{\mathcal{M}, \mathbf{X}}$ overall.

\textit{\textbf{Remark:} } It may so happen that the responses to two intent-aligned prompts $x_i$ and $x_j$ are semantically equivalent but involve very different word choices and still $\mathds{P}_{\mathcal{M}}(y_j|x_j)$ and $\mathds{P}_{\mathcal{M}}(y_j|x_i)$ can be significantly different. In fact, depending on the model and input prompts, this can happen even if $y_i$ and $y_j$ are the exact same strings! Therefore, in Section \ref{sec:approach_variance_in_confidence}, we also consider the variance in probabilities even if the strings are exactly the same. This is also a reason why average cosine similarity between the generated responses cannot by itself be employed as a prompt sensitivity metric.

\subsubsection{\metric\ and Variance in Confidence}
\label{sec:approach_variance_in_confidence}
Consider the case where all responses to the prompts in $\mathbf{X}$ are exactly the same, i.e., all $y_j$'s are the same. In such a case, should we call the model to be not sensitive at all? Upon a closer look, we can realize that the model would still be considered as sensitive if there is a notable variation in the likelihood of generating the response with a change in the input prompt.
As evident from Definition \ref{def:index}, our proposed index $\psi_{\mathcal{M}, \mathbf{X}}$ directly measures how divergent the log-likelihoods are for different intent-aligned prompts, thereby capturing the subtle nuances in response generation which contribute towards the sensitivity of the models.

\begin{table*}[h!]
    \centering
    \resizebox{\textwidth}{!}{
        \begin{tabular}{lccccccccc}
            \toprule
            \multirow{4}{*}{Model} & \multicolumn{4}{c}{MMLU-ZeroShot} & \multicolumn{4}{c}{Alpaca-ZeroShot} \\
            \cmidrule(lr){2-5}\cmidrule(lr){6-9}
            & \begin{tabular}{c} Spelling \\ Errors\end{tabular} & \begin{tabular}{c} Prompt \\ Templates \end{tabular} & Paraphrases & Mixture & \begin{tabular}{c} Spelling \\ Errors\end{tabular} & \begin{tabular}{c} Prompt \\ Templates \end{tabular} & Paraphrases & Mixture \\ 
            \midrule
            
            Llama-2-7b & $0.083_{\pm 0.073}$ & $1.12_{\pm 0.377}$ & $0.160_{\pm 0.160}$ & $0.821_{\pm 0.272}$ & $0.146_{\pm 0.115}$ & $0.202_{\pm 0.103}$ & $0.252_{\pm 0.192}$ & $0.271_{\pm 0.158}$  \\
            
            Llama-2-7b-chat & $0.082_{\pm 0.103}$ & $0.809_{\pm 0.283}$ & $0.135_{\pm 0.189}$ & $0.444_{\pm 0.258}$ & $0.246_{\pm 0.175}$ & $0.164_{\pm 0.139}$ & $0.66_{\pm 0.33}$ & $0.500_{\pm 0.229}$ \\
            
            Llama-3-8b & $0.086_{\pm 0.097}$ & $1.106_{\pm 0.612}$ & $0.11_{\pm 0.109}$ & $0.641_{\pm 0.383}$ & $0.123_{\pm 0.091}$ & $0.150_{\pm 0.107}$ & $0.249_{\pm 0.175}$ & $0.239_{\pm 0.136}$ \\
            
            Llama-3-8b-chat & $0.087_{\pm 0.09}$ & $1.048_{\pm 0.612}$ & $0.134_{\pm 0.126}$ & $0.650_{\pm 0.421}$ & $0.184_{\pm 0.152}$ & $0.15_{\pm 0.13}$ & $0.413_{\pm 0.259}$ & $0.357_{\pm 0.201}$ \\
            
            Mistral-7B & $0.065_{\pm 0.06}$ & $1.222_{\pm 0.571}$ & $0.108_{\pm 0.114}$ & $0.672_{\pm 0.303}$ & $0.18_{\pm 0.14}$ & $0.217_{\pm 0.148}$ & $0.242_{\pm 0.181}$ & $0.295_{\pm 0.181}$ \\
            
            Mistral-7B-Instruct & $0.105_{\pm 0.098}$ & $1.464_{\pm 0.528}$ & $0.126_{\pm 0.112}$ &
            $0.886_{\pm 0.328}$ & $0.195_{\pm 0.130}$ & $0.124_{\pm 0.069}$ & $0.296_{\pm 0.236}$ & $0.272_{\pm 0.152}$
            \\
            
            OLMo-7B-Base & $0.197_{\pm 0.207}$ & $1.672_{\pm 0.383}$ & $0.189_{\pm 0.164}$ & $1.134_{\pm 0.286}$ & $0.355_{\pm 0.305}$ & $0.369_{\pm 0.095}$ & $0.281_{\pm 0.199}$ & $0.448_{\pm 0.227}$ \\
            
            OLMo-7B-Instruct & $0.527_{\pm 0.485}$ & $1.499_{\pm 0.384}$ & $0.831_{\pm 0.595}$ & $1.413_{\pm 0.474}$ & $0.646_{\pm 0.378}$ & $0.192_{\pm 0.113}$ & $0.633_{\pm 0.382}$ & $0.62_{\pm 0.312}$ \\
            \hline
        \end{tabular}
    }
    \caption{\metric\ computed for 8 different models and 4 different prompt variation types on both MCQs (MMLU) and open-ended generation (Alpaca). The mixture variant consists of equal proportion of the other three variations.}
   \vspace{-2mm}
    \label{tab:main_table}
\end{table*}

\section{Experimental Setup}\label{sec:exp_setup}
To analyse the effectiveness of \metric\ in capturing various facets of sensitivity as described in Section~\ref{sec:approach} and to quantify and compare the prompt sensitivity of various LLMs using it, we experiment on Massive Multitask Language Understanding benchmark, or MMLU \cite{mmlu}, for classification tasks (posed as MCQ questions), and on Alpaca \cite{alpaca} for open-ended generation tasks. We also include Big Bench Hard, or BBH~\cite{suzgun2022challenging} as an additional dataset in Appendix~\ref{appendix:bbh}. MMLU contains about \(14,000\) prompt-response pairs from \(57\) different domains. For open-ended generation task, we sample \(5,000\) questions from Alpaca.

We consider a total of eight LLMs from three families: LLaMA, Mistral, and OLMo. These models include LLaMA-2 7B (base and chat variants) \cite{touvron2023llama}, LLaMA-3 8B (base and instruct variants), Mistral 7B (base and instruct variants) \cite{jiang2023mistral} and OLMo 7B (base and instruct variants) \cite{groeneveld2024olmo}. All our experiments were run on 8 NVIDIA A100-SXM4-80GB GPUs.

For each prompt, we generate \(60\) variants such that they are {\em intent-aligned} with each other and also with the original prompt. These 60 variants are composed of equal proportion of three types of variations --- introduction of minor spelling errors, alteration of the prompt template and paraphrasing/re-wording of the prompt. To study the effect of variation type on sensitivity, each prompt variant is generated using one of these three possible alterations. The variation types are discussed in detail below:

\noindent\textbf{Spelling Errors:} To introduce spelling errors, we randomly select one, two, four or eight tokens from the question in the prompt and introduce one of four possible spelling errors: (i) \textit{Insertion}, in which a random letter is added within the token; (ii) \textit{Omission}, in which a letter at a randomly chosen position is deleted; (iii) \textit{Transposition}, in which two adjacent letters are swapped; or (iv) \textit{Substitution}, in which a letter at a randomly chosen position is replaced with one of its adjacent letters on the keyboard. The specific error to be applied to each token is chosen randomly.

The \(20\) variations for spelling errors are derived from the combinations of the number of tokens with errors (one, two, four, eight) and five different seeds to control randomness. These types of spelling errors are based on the study by \citet{spell}.

\noindent\textbf{Prompt Templates:} For altering the prompt template, we design \(20\) different templates based on the grammar defined by \citet{sclar2023quantifying}. These templates are designed to maintain the core meaning of the prompt while altering its structure. The prompt templates used in our study are listed in Appendix \ref{appendix:templates}.

\noindent\textbf{Paraphrases:}
To create paraphrased variations of the prompts, we use GPT-3.5-Turbo to generate \(20\) paraphrases for each original prompt (The full dataset of generated paraphrases is \href{https://github.com/kowndinya-renduchintala/POSIX}{open-sourced} however a few examples of generated paraphrases are also present in Appendix~\ref{appendix:paraphrase_examples} for reference). These paraphrases are such that they rephrase the question while preserving its original intent and meaning.\\

\noindent For each prompt, we compute \metric\ for each type of variation, based on Definition \ref{def:index}, using an intent-aligned prompt set that includes the original prompt and its \(20\) corresponding variants. Additionally, for the \textbf{Mixture} category, we calculate \metric\ using a set of \(21\) prompt variants sampled uniformly from the three variation categories.

For experiments on MMLU, we generate five tokens as output, as the tasks are of multiple-choice question-answering format; whereas for the open-ended questions in Alpaca, 30 tokens are generated. \metric\ is, however, comparable for arbitrary response lengths as it is normalized by the number of tokens in the response. 

\begin{figure*}[tbh!]
    \centering
    \begin{subfigure}[b]{3.1in}
        \centering
        \includegraphics[width=3.48in]{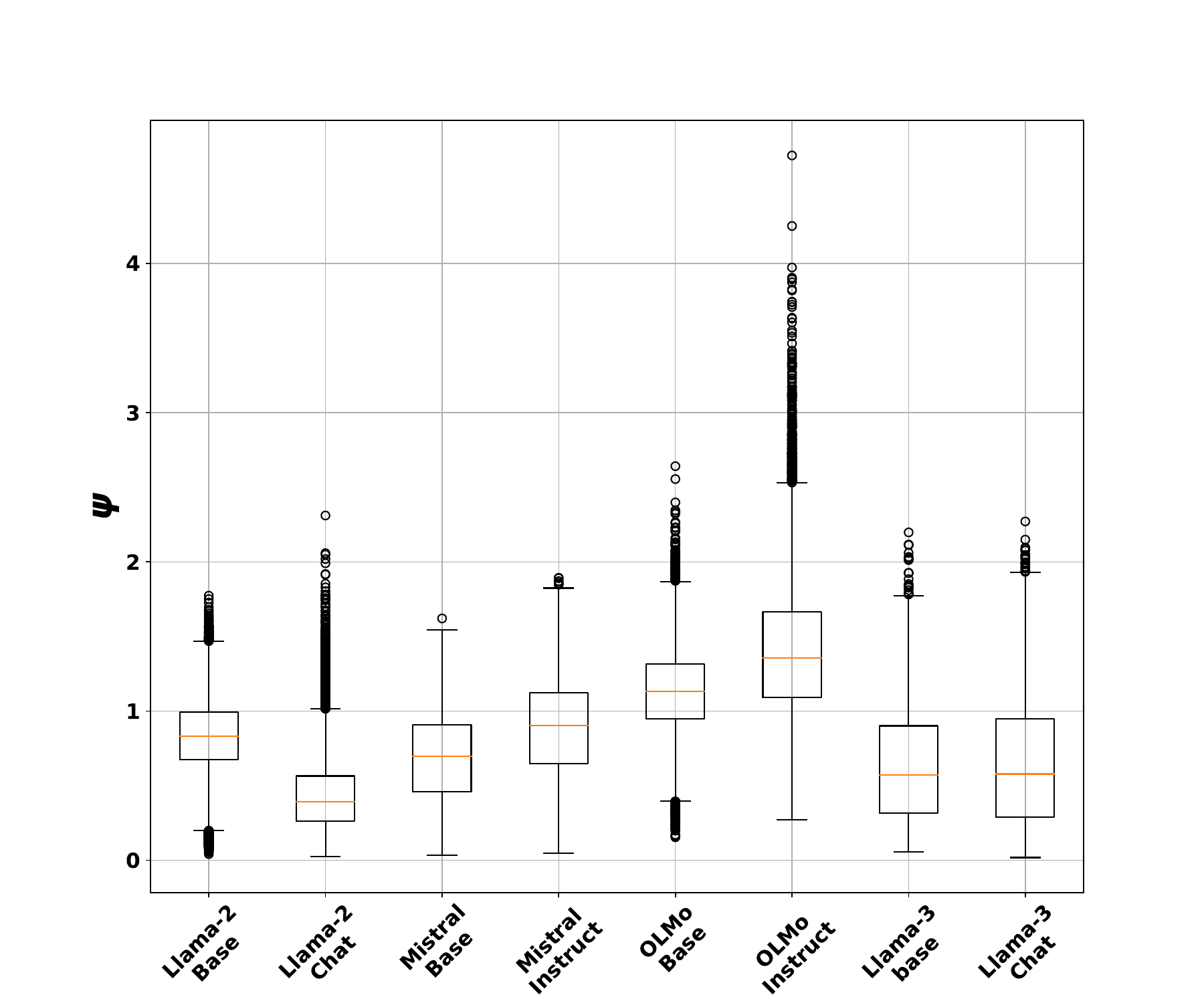}
        \caption{MMLU (MCQs)}
    \end{subfigure}
    \begin{subfigure}[b]{3.1in}
        \includegraphics[width=3.48in]{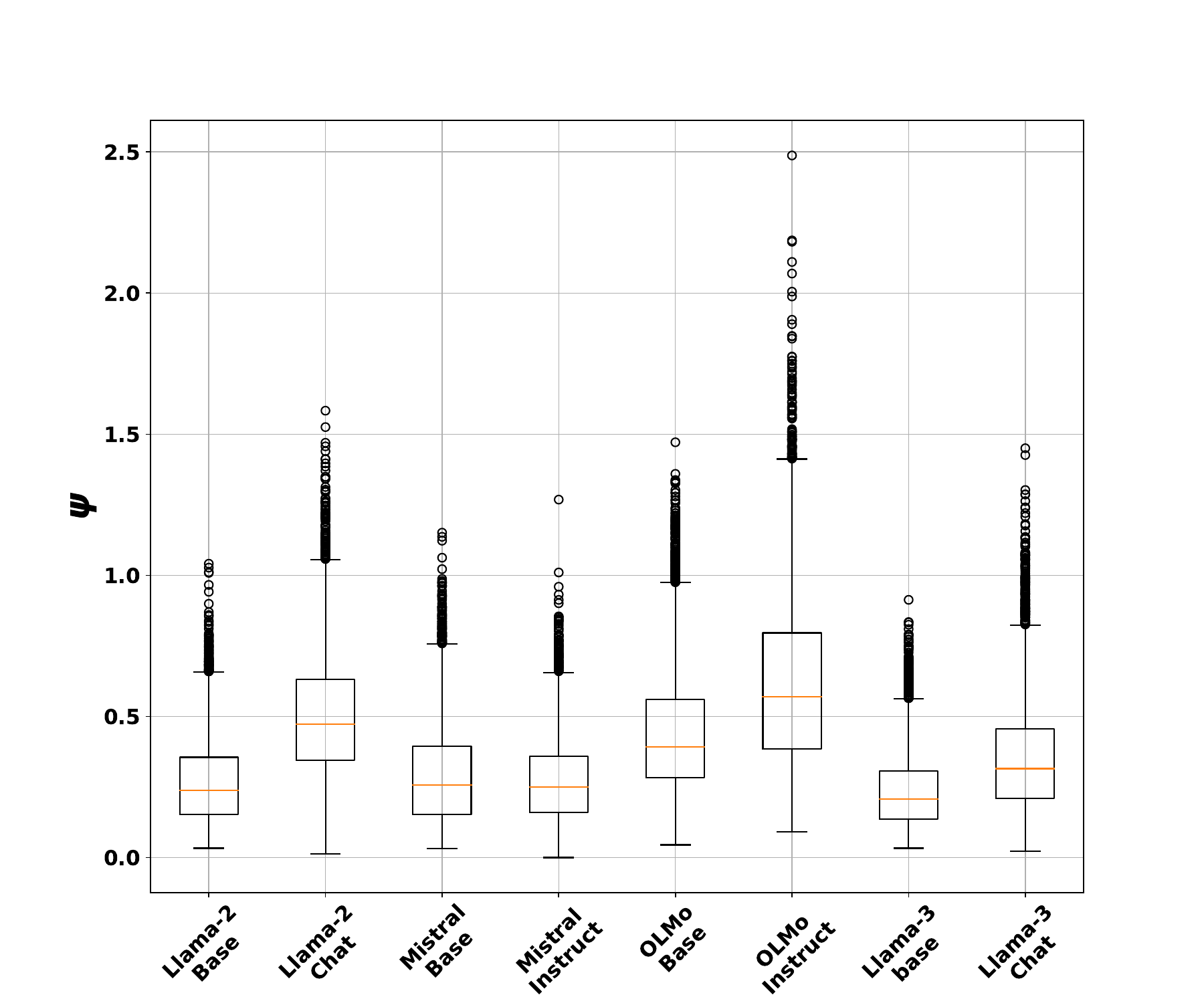}
        \caption{Alpaca (Open-ended generation)}
    \end{subfigure}
    \caption{Box plots depicting the distribution of $\psi_{\mathcal{M}, \mathbf{X}}$ for different instances of $\mathcal{M}$. The first plot corresponds to $\mathbf{X}$'s from MMLU dataset (MCQs) and the second plot corresponds to $\mathbf{X}$'s from the Alpaca dataset (open-ended generation).}
    \label{fig:chi_violin_plots}
\end{figure*}

\section{Results and Analysis}
\label{sec:results}

\subsection{Evaluating the Efficacy of \metric}\label{sec:goodness}
We now empirically investigate if \metric\ incorporates the four factors described in Section~\ref{sec:approach} by looking at correlation plots of \metric\ with those factors. For the plots, we combine data from all types of prompt variations and all models listed in Section~\ref{sec:exp_setup}. Additionally, for computing the cosine similarity between generated responses, we use an off-the-shelf Sentence Transformer model (GTE-large). Figure~\ref{fig:efficacy_mmlu} shows the correlation plots between \metric\ and each of the four properties described in Section~\ref{sec:approach}. For response distribution entropy, average cosine similarity and the log-probability variance, to observe the trend, we first bin the x-axes and corresponding average $\psi$ of the bins are plotted. The number of unique responses, the response distribution entropy and the log-probability variance (in case all responses are identical) --- all have positive correlation with \metric\ and the average cosine similarity between the responses is negatively correlated with \metric. Thus, the plots serve as an empirical validation for the fact that \metric\ incorporates each of the factors described in Section~\ref{sec:approach}. Please refer to Appendix~\ref{app:efficacy_figure_alpaca} for the correlation plots of open-ended generation (Alpaca).

\begin{table*}[ht!]
\small
    \centering
    \resizebox{\textwidth}{!}{
        \begin{tabular}{lcccccc}
            \toprule
            n\_shot & Variation Type & Llama-2-7b & Llama-2-7b-chat & Mistral-7B & Mistral-7B-Instruct \\
            \midrule
            \multirow{3}{*}{0-shot} & Spelling Errors & $0.083_{\pm 0.073}$ & $0.082_{\pm 0.103}$ & $0.065_{\pm 0.06}$ & $0.105_{\pm 0.098}$  \\
            & Prompt Templates & $1.12_{\pm 0.377}$& $0.809_{\pm 0.283}$ & $1.222_{\pm 0.571}$& $1.464_{\pm 0.0.528}$ \\
            & Paraphrases & $0.16_{\pm 0.16}$ & $0.135_{\pm 0.189}$ & $0.108_{\pm 0.115}$ & $0.126_{\pm 0.112}$ \\
            \hline
            \multirow{3}{*}{1-shot} & Spelling Errors & $0.026_{\pm 0.021}$ & $0.048_{\pm 0.066}$ & $0.042_{\pm 0.039}$ & $0.087_{\pm 0.065}$ \\
            & Prompt Templates & $0.513_{\pm 0.347}$ & $0.357_{\pm 0.169}$& $0.2_{\pm 0.244}$ & $1.387_{\pm 0.707}$ \\
            & Paraphrases & $0.035_{\pm 0.031}$ & $0.064_{\pm 0.0.07}$ & $0.046_{\pm 0.045}$ & $0.085_{\pm 0.081}$ \\
            \hline
            \multirow{3}{*}{2-shot} & Spelling Errors & $0.027_{\pm 0.024}$ & $0.049_{\pm 0.07}$ & $0.042_{\pm 0.041}$ & $0.085_{\pm 0.072}$ \\
            & Prompt Templates & $0.482_{\pm 0.38}$& $0.272_{\pm 0.117}$& $0.225_{\pm 0.247}$ & $1.128_{\pm 0.773}$ \\
            & Paraphrases & $0.036_{\pm 0.035}$ & $0.065_{\pm 0.074}$ & $0.047_{\pm 0.047}$ & $0.085_{\pm 0.09}$ & \\
            \hline
            \multirow{3}{*}{3-shot} & Spelling Errors & $0.028_{\pm 0.024}$& $0.051_{\pm 0.073}$ & $0.043_{\pm 0.041}$ & $0.088_{\pm 0.073}$ \\
            & Prompt Templates & $0.554_{\pm 0.433}$ & $0.249_{\pm 0.091}$ & $0.23_{\pm 0.247}$ & $1.101_{\pm 0.775}$ \\
            & Paraphrases & $0.039_{\pm 0.039}$ & $0.068_{\pm 0.077}$ & $0.047_{\pm 0.047}$& $0.086_{\pm 0.0.98}$ \\
            \hline
        \end{tabular}
    }
    \caption{\metric\ computed for Llama-2 and Mistral models on the MMLU dataset in few-shot settings.}
    \label{tab:fewshot_table}
\end{table*}    

\subsection{Effect of Instruction Tuning on Sensitivity} 
Table \ref{tab:main_table} presents the \metric\ values of various models obtained for different variation types on MMLU and Alpaca (Please refer to Appendix ~\ref{appendix:bbh} for results on BBH). We observe that chat or instruct versions of the models are generally less sensitive than the corresponding base models in the case of template variations on MMLU, with Mistral being the only exception. However, in the other categories of variations, the instruct versions tend to be more sensitive than their base models. Especially for open-ended generation tasks, i.e., on Alpaca, the higher sensitivity of instruct versions is even more pronounced. Note that this implies that instruction tuning does not necessarily improve model sensitivity. Figure \ref{fig:chi_violin_plots} shows the distribution of \metric\ values of all models for the \textit{Mixture} of variation types (please refer to Appendix~\ref{appendix:box_plots} for the other variation types). We clearly observe the instruct models to have higher sensitivity than the base ones, more so on Alpaca. Furthermore, the Mistral models seem to have the least divergence in sensitivity between the base and instruct variants whereas the OLMo models have the most disparity.

As the base models undergo both instruction tuning and alignment on human preferences to obtain the instruct versions, the above observations are a cumulative effect of instruction tuning and alignment procedures. To disentangle their consequences and study the effect of only instruction tuning on sensitivity, we separately fine-tune LLaMA-2 7B and Mistral 7B base models on the entire FLAN dataset \cite{weifinetuned}. We call these fine-tuned models Llama-2-7b-FLAN and Mistral-7B-FLAN, respectively. Their \metric\ values are reported in Table \ref{tab:flan_table}.
In most cases, the chat version is better than FLAN-only models in terms of sensitivity, except in case of prompt template for MMLU, where FLAN-only models significantly outperform the chat versions. We hypothesize that this is due to the nature of the FLAN dataset, which contains a huge focus on MCQs and various prompt templates. This might have given the FLAN-only models an edge over the chat versions.

\begin{table}[]
    \centering
    \scalebox{0.68}{
        \begin{tabular}{lccc}
            \toprule
            Dataset & Variation Type & Llama-2-7b-FLAN & Mistral-7B-FLAN \\
            \midrule
            \multirow{3}{*}{MMLU} & Spelling Errors & $0.113_{\pm 0.116}$ & $0.14_{\pm 0.143}$  \\
            & Prompt Templates & $0.229_{\pm 0.0.169}$& $0.668_{\pm 0.614}$ \\
            & Paraphrases & $0.163_{\pm 0.136}$ & $0.187_{\pm 0.162}$ \\
            
            \hline
            \multirow{3}{*}{Alpaca} & Spelling Errors & $0.317_{\pm 0.177}$& $0.284_{\pm 0.177}$ \\
            & Prompt Templates & $0.166_{\pm 0.129}$ & $0.163_{\pm 0.145}$  \\
            & Paraphrases & $0.267_{\pm 0.192}$ & $0.278_{\pm 0.2}$ \\
            \hline
        \end{tabular}
    }
    \caption{\metric\ computed for Llama-2-7b-FLAN and Mistral-7B-FLAN (unlike the chat versions, these are only instruction-tuned on the FLAN dataset without any further RLHF).}
    \vspace{-4mm}
    \label{tab:flan_table}
\end{table}

\subsection{Impact of Model Scale on Sensitivity}

\begin{figure}
    \centering
\includegraphics[width=\linewidth]{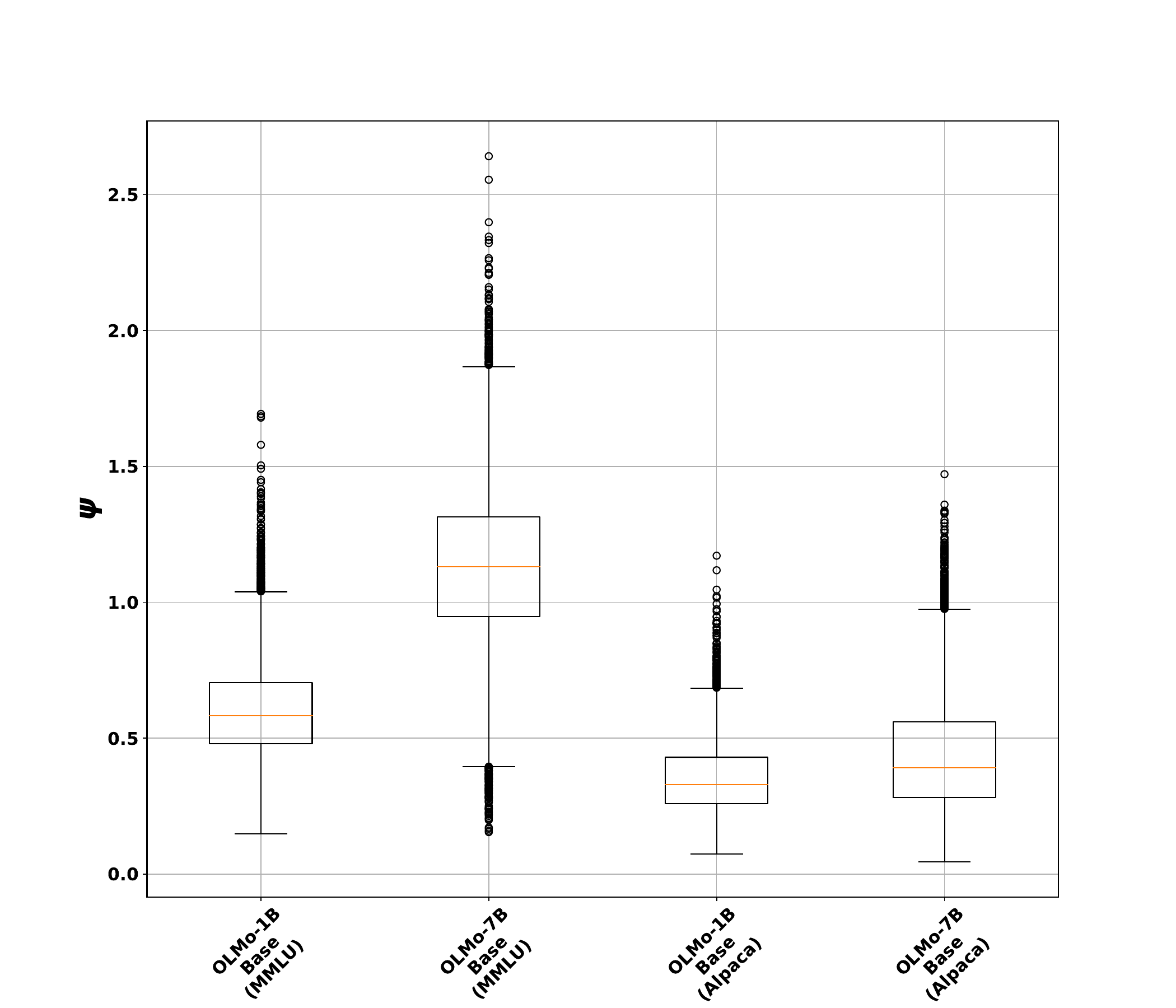}
    \caption{Box plots depicting distribution of $\psi_{\mathcal{M}, \mathbf{X}}$ for two differently sized OLMo models (1B and 7B).}
    \label{fig:model_scale_figure}
\end{figure}

\begin{figure*}
    \centering
    \begin{subfigure}[b]{3.1in}
        \centering
        \includegraphics[width=3.48in]{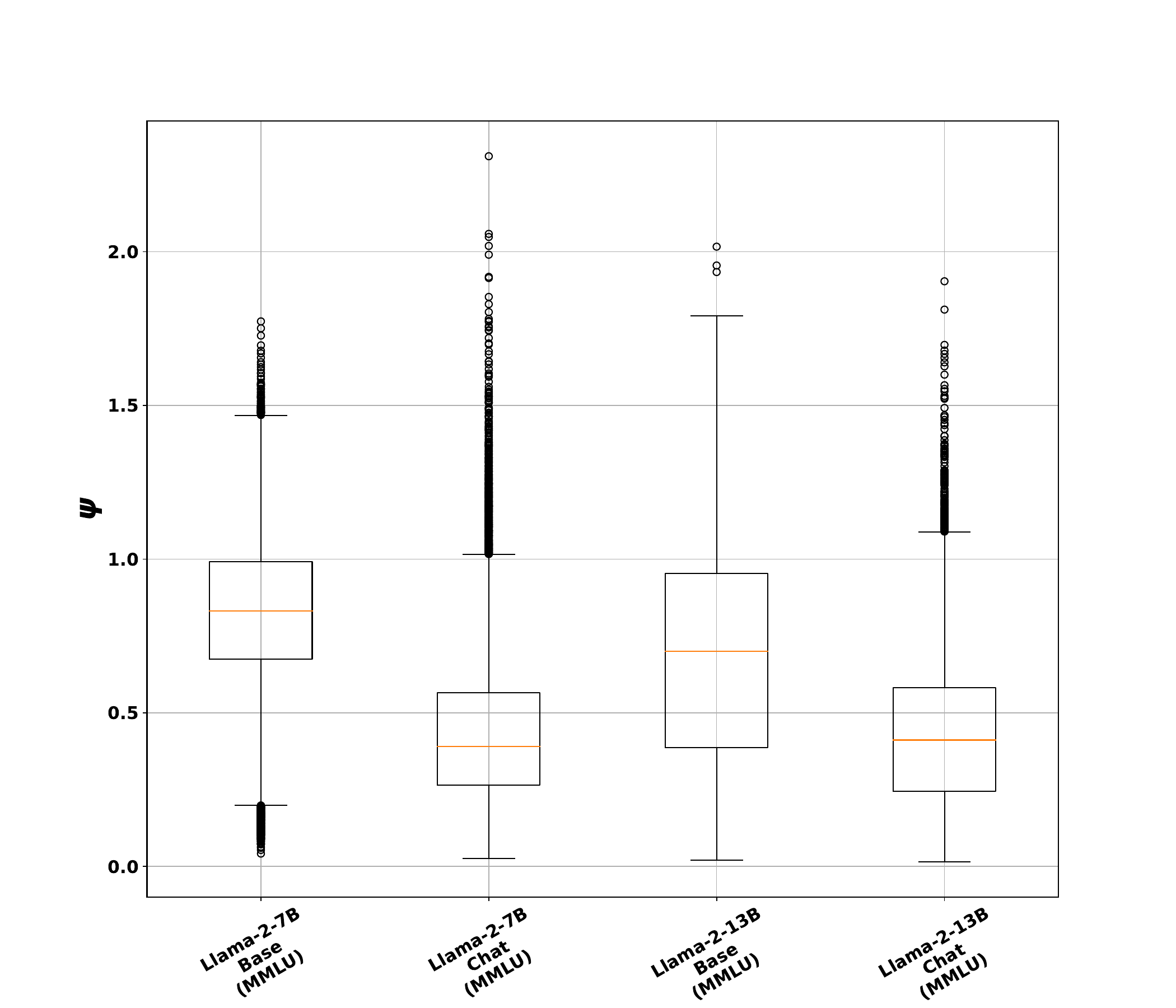}
        \caption{MMLU (MCQs)}
    \end{subfigure}
    \begin{subfigure}[b]{3.1in}
        \centering
        \includegraphics[width=3.48in]{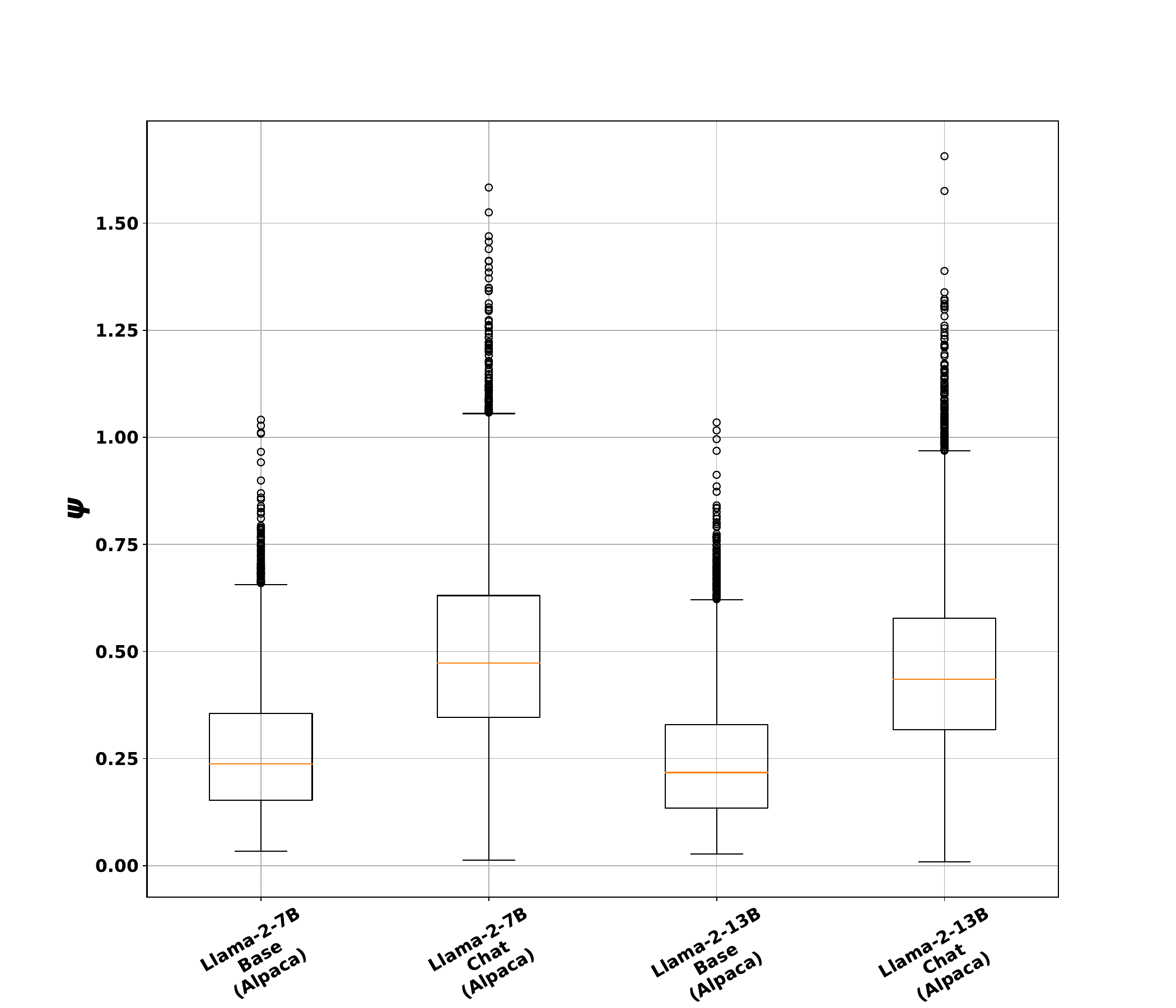}
        \caption{Alpaca (Open-ended generation)}
    \end{subfigure}

    \caption{Box plots depicting distribution of $\psi_{\mathcal{M}, \mathbf{X}}$ for two differently sized Llama-2 models (7B and 13B).}
     \vspace{-3mm}
    \label{fig:llama_2_model_scale_figure}
\end{figure*}

To study the effect of model scale on prompt sensitivity, we experiment with 1B and 7B variants of OLMo as well as 7B and 13B variants of Llama-2. Figure \ref{fig:model_scale_figure} and Figure \ref{fig:llama_2_model_scale_figure} show the variations in the value of \metric\ with model scale for prompt variants from \textit{Mixture} of variation types, on both MMLU and Alpaca (Please refer to Appendix ~\ref{appendix:bbh} for results on BBH).  We observe that the 7B model is significantly more sensitive compared to the 1B model on the MMLU dataset; however, they are comparable in the case of the Alpaca dataset, in the case of OLMo. Similarly, even in the case of Llama-2, a 13B model is not guaranteed to always have lesser prompt sensitivity than a 7B model. This only re-emphasizes the fact that accuracy and sensitivity are two separate aspects --- and higher accuracy does not necessarily imply better sensitivity and vice-versa. The \metric\ values of OLMo-1B for all variation types are reported in Table \ref{tab:olmo1b} of Appendix \ref{appendix:olmo1b}, and that of Llama-2-13B (base and chat) models are reported in Table \ref{tab:llama-2-13b-MMLU} (for MMLU) and Table \ref{tab:llama-2-13b-Alpaca} (for Alpaca) of Appendix \ref{appendix:llama13b}.

\subsection{Few-shot Exemplars and Sensitivity} 

Table~\ref{tab:fewshot_table} consists of \metric\ values computed for the few-shot setting in the case of the MMLU dataset. The key finding is that adding few-shot exemplars, even if it just a single example can significantly boost the robustness of LLMs towards variations in prompts. Although, adding even more few-shot examples might yield diminishing gains, i.e., when compared to the value that a single example adds, the additional value of a second or third few-shot exemplar is not that much --- prompt sensitivity either remains about the same or slightly decreases.

\subsection{Impact of Various Variation Categories} 
From Table~\ref{tab:main_table}, it can be observed that prompt template is the most sensitive variation type in the case of MCQs, and paraphrases are almost always the most sensitive variation type in the case of Alpaca (OLMo being the only exception). Moreover, the sensitivity of the prompt template is more sensitive than any other perturbation in the case of Alpaca, thus making MCQ more sensitive overall (please refer to the \textit{mixture} column in Table \ref{tab:main_table}). Based on this prompt sensitivity analysis, we could offer some insights while performing prompt engineering as well --- for MCQs, it is better to invest efforts in getting the proper prompt template while for open-ended questions, it is crucial to re-phrase the query properly.
\section{Conclusions}
We introduced \metric\ - a novel prompt sensitivity index, as a reliable measure of sensitivity of LLMs towards intent-preserving variations in prompts such as spelling errors, prompt templates, and alterations in the wording. We presented thorough empirical analysis for the efficacy of \metric\ in capturing prompt sensitivity and subsequently used it to measure and compare multiple open-source LLMs, revealing some interesting observations such as prompt template is the most sensitive variant type for MCQ tasks and paraphrasing is the most sensitive variant type for open-ended generation tasks, and also that parameter count or instruction tuning do not necessarily decrease prompt sensitivity of the models. These findings highlight the nuanced behaviour of LLMs towards prompt variations, underscoring the importance of considering prompt sensitivity index for their holistic evaluation.

\section*{Limitations}
While \metric\ has its own advantages, like the ability to work across different kinds of prompt variations and tasks, including open-ended generation with arbitrary response lengths, one of the main limitations of \metric\ is its computational complexity. \metric\ needs $\mathcal{O}(MN^2)$ log-likelihood comparisons if $M$ is the total number of prompts in a dataset under consideration and $N$ is the number of variations per prompt. Nevertheless, \metric\ is very effective in incorporating various facets of prompt sensitivity.

\section*{Ethical Considerations}
Since we use open-source large language models and open-source datasets like MMLU and Alpaca, our work encompasses all the corresponding considerations of those works. Although, our method would be expected to largely benefit the community by providing a reliable way to evaluate sensitivity of large language models towards variations in prompts. While attempting to paraphrase the prompts in MMLU using GPT-3.5-Turbo, quite a few prompts have been flagged as either violent or biased, etc. Most of them were from the \textit{moral\_scenarios} split of MMLU. We made sure to remove these from our analyses.

\section*{Acknowledgments}
Tanmoy Chakraborty acknowledges the financial support of Adobe Faculty Award. 
\bibliography{custom}

\begin{thebibliography}{42}
\providecommand{\natexlab}[1]{#1}

\bibitem[{AI@Meta(2024)}]{llama3modelcard}
AI@Meta. 2024.
\newblock \href {https://github.com/meta-llama/llama3/blob/main/MODEL_CARD.md} {Llama 3 model card}.

\bibitem[{Almazrouei et~al.(2023)Almazrouei, Alobeidli, Alshamsi, Cappelli, Cojocaru, Debbah, Goffinet, Hesslow, Launay, Malartic et~al.}]{almazrouei2023falcon}
Ebtesam Almazrouei, Hamza Alobeidli, Abdulaziz Alshamsi, Alessandro Cappelli, Ruxandra Cojocaru, M{\'e}rouane Debbah, {\'E}tienne Goffinet, Daniel Hesslow, Julien Launay, Quentin Malartic, et~al. 2023.
\newblock \href {https://arxiv.org/abs/2311.16867} {The falcon series of open language models}.
\newblock \emph{ArXiv preprint}, abs/2311.16867.

\bibitem[{Arora et~al.(2022)Arora, Narayan, Chen, Orr, Guha, Bhatia, Chami, Sala, and Ré}]{arora2022ask}
Simran Arora, Avanika Narayan, Mayee~F. Chen, Laurel Orr, Neel Guha, Kush Bhatia, Ines Chami, Frederic Sala, and Christopher Ré. 2022.
\newblock \href {https://arxiv.org/abs/2210.02441} {Ask me anything: A simple strategy for prompting language models}.
\newblock \emph{Preprint}, arXiv:2210.02441.

\bibitem[{Brooks et~al.(1993)Brooks, Gorman, and Kendall}]{spell}
Greg Brooks, Tom Gorman, and Lesley Kendall. 1993.
\newblock Spelling it out: the spelling abilities of 11- and 15-year-olds.
\newblock Technical report, National Foundation for Educational Research (NFER), Slough.

\bibitem[{Brown et~al.(2020)Brown, Mann, Ryder, Subbiah, Kaplan, Dhariwal, Neelakantan, Shyam, Sastry, Askell, Agarwal, Herbert{-}Voss, Krueger, Henighan, Child, Ramesh, Ziegler, Wu, Winter, Hesse, Chen, Sigler, Litwin, Gray, Chess, Clark, Berner, McCandlish, Radford, Sutskever, and Amodei}]{brown2020language}
Tom~B. Brown, Benjamin Mann, Nick Ryder, Melanie Subbiah, Jared Kaplan, Prafulla Dhariwal, Arvind Neelakantan, Pranav Shyam, Girish Sastry, Amanda Askell, Sandhini Agarwal, Ariel Herbert{-}Voss, Gretchen Krueger, Tom Henighan, Rewon Child, Aditya Ramesh, Daniel~M. Ziegler, Jeffrey Wu, Clemens Winter, Christopher Hesse, Mark Chen, Eric Sigler, Mateusz Litwin, Scott Gray, Benjamin Chess, Jack Clark, Christopher Berner, Sam McCandlish, Alec Radford, Ilya Sutskever, and Dario Amodei. 2020.
\newblock \href {https://proceedings.neurips.cc/paper/2020/hash/1457c0d6bfcb4967418bfb8ac142f64a-Abstract.html} {Language models are few-shot learners}.
\newblock In \emph{Advances in Neural Information Processing Systems 33: Annual Conference on Neural Information Processing Systems 2020, NeurIPS 2020, December 6-12, 2020, virtual}.

\bibitem[{Deng et~al.(2022)Deng, Wang, Hsieh, Wang, Guo, Shu, Song, Xing, and Hu}]{deng-etal-2022-rlprompt}
Mingkai Deng, Jianyu Wang, Cheng-Ping Hsieh, Yihan Wang, Han Guo, Tianmin Shu, Meng Song, Eric Xing, and Zhiting Hu. 2022.
\newblock \href {https://aclanthology.org/2022.emnlp-main.222} {{RLP}rompt: Optimizing discrete text prompts with reinforcement learning}.
\newblock In \emph{Proceedings of the 2022 Conference on Empirical Methods in Natural Language Processing}, pages 3369--3391, Abu Dhabi, United Arab Emirates. Association for Computational Linguistics.

\bibitem[{Efrat et~al.(2023)Efrat, Honovich, and Levy}]{efrat-etal-2023-lmentry}
Avia Efrat, Or~Honovich, and Omer Levy. 2023.
\newblock \href {https://doi.org/10.18653/v1/2023.findings-acl.666} {{LM}entry: A language model benchmark of elementary language tasks}.
\newblock In \emph{Findings of the Association for Computational Linguistics: ACL 2023}, pages 10476--10501, Toronto, Canada. Association for Computational Linguistics.

\bibitem[{Freeman(1965)}]{freeman1965elementary}
Linton~C Freeman. 1965.
\newblock Elementary applied statistics: for students in behavioral science.
\newblock \emph{(No Title)}.

\bibitem[{Gao et~al.(2021)Gao, Fisch, and Chen}]{gao-etal-2021-making}
Tianyu Gao, Adam Fisch, and Danqi Chen. 2021.
\newblock \href {https://doi.org/10.18653/v1/2021.acl-long.295} {Making pre-trained language models better few-shot learners}.
\newblock In \emph{Proceedings of the 59th Annual Meeting of the Association for Computational Linguistics and the 11th International Joint Conference on Natural Language Processing (Volume 1: Long Papers)}, pages 3816--3830, Online. Association for Computational Linguistics.

\bibitem[{Groeneveld et~al.(2024)Groeneveld, Beltagy, Walsh, Bhagia, Kinney, Tafjord, Jha, Ivison, Magnusson, Wang, Arora, Atkinson, Authur, Chandu, Cohan, Dumas, Elazar, Gu, Hessel, Khot, Merrill, Morrison, Muennighoff, Naik, Nam, Peters, Pyatkin, Ravichander, Schwenk, Shah, Smith, Strubell, Subramani, Wortsman, Dasigi, Lambert, Richardson, Zettlemoyer, Dodge, Lo, Soldaini, Smith, and Hajishirzi}]{groeneveld2024olmo}
Dirk Groeneveld, Iz~Beltagy, Pete Walsh, Akshita Bhagia, Rodney Kinney, Oyvind Tafjord, Ananya~Harsh Jha, Hamish Ivison, Ian Magnusson, Yizhong Wang, Shane Arora, David Atkinson, Russell Authur, Khyathi~Raghavi Chandu, Arman Cohan, Jennifer Dumas, Yanai Elazar, Yuling Gu, Jack Hessel, Tushar Khot, William Merrill, Jacob Morrison, Niklas Muennighoff, Aakanksha Naik, Crystal Nam, Matthew~E. Peters, Valentina Pyatkin, Abhilasha Ravichander, Dustin Schwenk, Saurabh Shah, Will Smith, Emma Strubell, Nishant Subramani, Mitchell Wortsman, Pradeep Dasigi, Nathan Lambert, Kyle Richardson, Luke Zettlemoyer, Jesse Dodge, Kyle Lo, Luca Soldaini, Noah~A. Smith, and Hannaneh Hajishirzi. 2024.
\newblock \href {https://arxiv.org/abs/2402.00838} {Olmo: Accelerating the science of language models}.
\newblock \emph{Preprint}, arXiv:2402.00838.

\bibitem[{Gu et~al.(2023)Gu, Zhao, Xu, Nie, Mei, and Yin}]{gu-etal-2023-robustness}
Jiasheng Gu, Hongyu Zhao, Hanzi Xu, Liangyu Nie, Hongyuan Mei, and Wenpeng Yin. 2023.
\newblock \href {https://doi.org/10.18653/v1/2023.findings-acl.875} {Robustness of learning from task instructions}.
\newblock In \emph{Findings of the Association for Computational Linguistics: ACL 2023}, pages 13935--13948, Toronto, Canada. Association for Computational Linguistics.

\bibitem[{Hendrycks et~al.(2021)Hendrycks, Burns, Basart, Zou, Mazeika, Song, and Steinhardt}]{mmlu}
Dan Hendrycks, Collin Burns, Steven Basart, Andy Zou, Mantas Mazeika, Dawn Song, and Jacob Steinhardt. 2021.
\newblock \href {https://openreview.net/forum?id=d7KBjmI3GmQ} {Measuring massive multitask language understanding}.
\newblock In \emph{9th International Conference on Learning Representations, {ICLR} 2021, Virtual Event, Austria, May 3-7, 2021}. OpenReview.net.

\bibitem[{Jiang et~al.(2023)Jiang, Sablayrolles, Mensch, Bamford, Chaplot, de~las Casas, Bressand, Lengyel, Lample, Saulnier, Lavaud, Lachaux, Stock, Scao, Lavril, Wang, Lacroix, and Sayed}]{jiang2023mistral}
Albert~Q. Jiang, Alexandre Sablayrolles, Arthur Mensch, Chris Bamford, Devendra~Singh Chaplot, Diego de~las Casas, Florian Bressand, Gianna Lengyel, Guillaume Lample, Lucile Saulnier, Lélio~Renard Lavaud, Marie-Anne Lachaux, Pierre Stock, Teven~Le Scao, Thibaut Lavril, Thomas Wang, Timothée Lacroix, and William~El Sayed. 2023.
\newblock \href {https://arxiv.org/abs/2310.06825} {Mistral 7b}.
\newblock \emph{Preprint}, arXiv:2310.06825.

\bibitem[{Kojima et~al.(2022)Kojima, Gu, Reid, Matsuo, and Iwasawa}]{kojima2022large}
Takeshi Kojima, Shixiang~Shane Gu, Machel Reid, Yutaka Matsuo, and Yusuke Iwasawa. 2022.
\newblock Large language models are zero-shot reasoners.
\newblock \emph{Advances in neural information processing systems}, 35:22199--22213.

\bibitem[{Kwiatkowski et~al.(2019)Kwiatkowski, Palomaki, Redfield, Collins, Parikh, Alberti, Epstein, Polosukhin, Devlin, Lee, Toutanova, Jones, Kelcey, Chang, Dai, Uszkoreit, Le, and Petrov}]{kwiatkowski-etal-2019-natural}
Tom Kwiatkowski, Jennimaria Palomaki, Olivia Redfield, Michael Collins, Ankur Parikh, Chris Alberti, Danielle Epstein, Illia Polosukhin, Jacob Devlin, Kenton Lee, Kristina Toutanova, Llion Jones, Matthew Kelcey, Ming-Wei Chang, Andrew~M. Dai, Jakob Uszkoreit, Quoc Le, and Slav Petrov. 2019.
\newblock \href {https://doi.org/10.1162/tacl_a_00276} {Natural questions: A benchmark for question answering research}.
\newblock \emph{Transactions of the Association for Computational Linguistics}, 7:452--466.

\bibitem[{Leidinger et~al.(2023)Leidinger, van Rooij, and Shutova}]{leidinger2023language}
Alina Leidinger, Robert van Rooij, and Ekaterina Shutova. 2023.
\newblock \href {https://arxiv.org/abs/2311.01967} {The language of prompting: What linguistic properties make a prompt successful?}
\newblock \emph{Preprint}, arXiv:2311.01967.

\bibitem[{Liang et~al.(2023)Liang, Bommasani, Lee, Tsipras, Soylu, Yasunaga, Zhang, Narayanan, Wu, Kumar, Newman, Yuan, Yan, Zhang, Cosgrove, Manning, Ré, Acosta-Navas, Hudson, Zelikman, Durmus, Ladhak, Rong, Ren, Yao, Wang, Santhanam, Orr, Zheng, Yuksekgonul, Suzgun, Kim, Guha, Chatterji, Khattab, Henderson, Huang, Chi, Xie, Santurkar, Ganguli, Hashimoto, Icard, Zhang, Chaudhary, Wang, Li, Mai, Zhang, and Koreeda}]{liang2023holistic}
Percy Liang, Rishi Bommasani, Tony Lee, Dimitris Tsipras, Dilara Soylu, Michihiro Yasunaga, Yian Zhang, Deepak Narayanan, Yuhuai Wu, Ananya Kumar, Benjamin Newman, Binhang Yuan, Bobby Yan, Ce~Zhang, Christian Cosgrove, Christopher~D. Manning, Christopher Ré, Diana Acosta-Navas, Drew~A. Hudson, Eric Zelikman, Esin Durmus, Faisal Ladhak, Frieda Rong, Hongyu Ren, Huaxiu Yao, Jue Wang, Keshav Santhanam, Laurel Orr, Lucia Zheng, Mert Yuksekgonul, Mirac Suzgun, Nathan Kim, Neel Guha, Niladri Chatterji, Omar Khattab, Peter Henderson, Qian Huang, Ryan Chi, Sang~Michael Xie, Shibani Santurkar, Surya Ganguli, Tatsunori Hashimoto, Thomas Icard, Tianyi Zhang, Vishrav Chaudhary, William Wang, Xuechen Li, Yifan Mai, Yuhui Zhang, and Yuta Koreeda. 2023.
\newblock \href {https://arxiv.org/abs/2211.09110} {Holistic evaluation of language models}.
\newblock \emph{Preprint}, arXiv:2211.09110.

\bibitem[{Liu et~al.(2022)Liu, Shen, Zhang, Dolan, Carin, and Chen}]{liu-etal-2022-makes}
Jiachang Liu, Dinghan Shen, Yizhe Zhang, Bill Dolan, Lawrence Carin, and Weizhu Chen. 2022.
\newblock \href {https://doi.org/10.18653/v1/2022.deelio-1.10} {What makes good in-context examples for {GPT}-3?}
\newblock In \emph{Proceedings of Deep Learning Inside Out (DeeLIO 2022): The 3rd Workshop on Knowledge Extraction and Integration for Deep Learning Architectures}, pages 100--114, Dublin, Ireland and Online. Association for Computational Linguistics.

\bibitem[{Liu et~al.(2023)Liu, Yuan, Fu, Jiang, Hayashi, and Neubig}]{liu2023pre}
Pengfei Liu, Weizhe Yuan, Jinlan Fu, Zhengbao Jiang, Hiroaki Hayashi, and Graham Neubig. 2023.
\newblock Pre-train, prompt, and predict: A systematic survey of prompting methods in natural language processing.
\newblock \emph{ACM Computing Surveys}, 55(9):1--35.

\bibitem[{Lu et~al.(2023)Lu, Schuff, and Gurevych}]{lu2023prompts}
Sheng Lu, Hendrik Schuff, and Iryna Gurevych. 2023.
\newblock \href {https://arxiv.org/abs/2311.07230} {How are prompts different in terms of sensitivity?}
\newblock \emph{Preprint}, arXiv:2311.07230.

\bibitem[{Lu et~al.(2022)Lu, Bartolo, Moore, Riedel, and Stenetorp}]{lu-etal-2022-fantastically}
Yao Lu, Max Bartolo, Alastair Moore, Sebastian Riedel, and Pontus Stenetorp. 2022.
\newblock \href {https://doi.org/10.18653/v1/2022.acl-long.556} {Fantastically ordered prompts and where to find them: Overcoming few-shot prompt order sensitivity}.
\newblock In \emph{Proceedings of the 60th Annual Meeting of the Association for Computational Linguistics (Volume 1: Long Papers)}, pages 8086--8098, Dublin, Ireland. Association for Computational Linguistics.

\bibitem[{Min et~al.(2022)Min, Lyu, Holtzman, Artetxe, Lewis, Hajishirzi, and Zettlemoyer}]{min-etal-2022-rethinking}
Sewon Min, Xinxi Lyu, Ari Holtzman, Mikel Artetxe, Mike Lewis, Hannaneh Hajishirzi, and Luke Zettlemoyer. 2022.
\newblock \href {https://aclanthology.org/2022.emnlp-main.759} {Rethinking the role of demonstrations: What makes in-context learning work?}
\newblock In \emph{Proceedings of the 2022 Conference on Empirical Methods in Natural Language Processing}, pages 11048--11064, Abu Dhabi, United Arab Emirates. Association for Computational Linguistics.

\bibitem[{Mizrahi et~al.(2024)Mizrahi, Kaplan, Malkin, Dror, Shahaf, and Stanovsky}]{mizrahi2024state}
Moran Mizrahi, Guy Kaplan, Dan Malkin, Rotem Dror, Dafna Shahaf, and Gabriel Stanovsky. 2024.
\newblock \href {https://arxiv.org/abs/2401.00595} {State of what art? a call for multi-prompt llm evaluation}.
\newblock \emph{Preprint}, arXiv:2401.00595.

\bibitem[{Polo et~al.(2024)Polo, Xu, Weber, Silva, Bhardwaj, Choshen, de~Oliveira, Sun, and Yurochkin}]{polo2024efficient}
Felipe~Maia Polo, Ronald Xu, Lucas Weber, Mírian Silva, Onkar Bhardwaj, Leshem Choshen, Allysson Flavio~Melo de~Oliveira, Yuekai Sun, and Mikhail Yurochkin. 2024.
\newblock \href {https://arxiv.org/abs/2405.17202} {Efficient multi-prompt evaluation of llms}.
\newblock \emph{Preprint}, arXiv:2405.17202.

\bibitem[{Radford et~al.(2019)Radford, Wu, Child, Luan, Amodei, and Sutskever}]{Radford2019LanguageMA}
Alec Radford, Jeff Wu, Rewon Child, David Luan, Dario Amodei, and Ilya Sutskever. 2019.
\newblock \href {https://api.semanticscholar.org/CorpusID:160025533} {Language models are unsupervised multitask learners}.

\bibitem[{Raffel et~al.(2020)Raffel, Shazeer, Roberts, Lee, Narang, Matena, Zhou, Li, and Liu}]{JMLR:v21:20-074}
Colin Raffel, Noam Shazeer, Adam Roberts, Katherine Lee, Sharan Narang, Michael Matena, Yanqi Zhou, Wei Li, and Peter~J. Liu. 2020.
\newblock \href {http://jmlr.org/papers/v21/20-074.html} {Exploring the limits of transfer learning with a unified text-to-text transformer}.
\newblock \emph{J. Mach. Learn. Res.}, 21:140:1--140:67.

\bibitem[{Reynolds and McDonell(2021)}]{10.1145/3411763.3451760}
Laria Reynolds and Kyle McDonell. 2021.
\newblock \href {https://doi.org/10.1145/3411763.3451760} {Prompt programming for large language models: Beyond the few-shot paradigm}.
\newblock In \emph{Extended Abstracts of the 2021 CHI Conference on Human Factors in Computing Systems}, CHI EA '21, New York, NY, USA. Association for Computing Machinery.

\bibitem[{Sclar et~al.(2023)Sclar, Choi, Tsvetkov, and Suhr}]{sclar2023quantifying}
Melanie Sclar, Yejin Choi, Yulia Tsvetkov, and Alane Suhr. 2023.
\newblock \href {https://arxiv.org/abs/2310.11324} {Quantifying language models' sensitivity to spurious features in prompt design or: How i learned to start worrying about prompt formatting}.
\newblock \emph{ArXiv preprint}, abs/2310.11324.

\bibitem[{Srivastava et~al.(2022)Srivastava, Rastogi, Rao, Shoeb, Abid, Fisch, Brown, Santoro, Gupta, Garriga-Alonso et~al.}]{srivastava2022beyond}
Aarohi Srivastava, Abhinav Rastogi, Abhishek Rao, Abu Awal~Md Shoeb, Abubakar Abid, Adam Fisch, Adam~R Brown, Adam Santoro, Aditya Gupta, Adri{\`a} Garriga-Alonso, et~al. 2022.
\newblock \href {https://arxiv.org/abs/2206.04615} {Beyond the imitation game: Quantifying and extrapolating the capabilities of language models}.
\newblock \emph{ArXiv preprint}, abs/2206.04615.

\bibitem[{Su et~al.(2022)Su, Kasai, Wu, Shi, Wang, Xin, Zhang, Ostendorf, Zettlemoyer, Smith, and Yu}]{su2022selective}
Hongjin Su, Jungo Kasai, Chen~Henry Wu, Weijia Shi, Tianlu Wang, Jiayi Xin, Rui Zhang, Mari Ostendorf, Luke Zettlemoyer, Noah~A. Smith, and Tao Yu. 2022.
\newblock \href {https://arxiv.org/abs/2209.01975} {Selective annotation makes language models better few-shot learners}.
\newblock \emph{Preprint}, arXiv:2209.01975.

\bibitem[{Sun et~al.(2024)Sun, Shaib, and Wallace}]{sun2024evaluating}
Jiuding Sun, Chantal Shaib, and Byron~C Wallace. 2024.
\newblock \href {https://openreview.net/forum?id=g9diuvxN6D} {Evaluating the zero-shot robustness of instruction-tuned language models}.
\newblock In \emph{The Twelfth International Conference on Learning Representations}.

\bibitem[{Suzgun et~al.(2022)Suzgun, Scales, Sch{\"a}rli, Gehrmann, Tay, Chung, Chowdhery, Le, Chi, Zhou et~al.}]{suzgun2022challenging}
Mirac Suzgun, Nathan Scales, Nathanael Sch{\"a}rli, Sebastian Gehrmann, Yi~Tay, Hyung~Won Chung, Aakanksha Chowdhery, Quoc~V Le, Ed~H Chi, Denny Zhou, et~al. 2022.
\newblock \href {https://arxiv.org/abs/2210.09261} {Challenging big-bench tasks and whether chain-of-thought can solve them}.
\newblock \emph{ArXiv preprint}, abs/2210.09261.

\bibitem[{Taori et~al.(2023)Taori, Gulrajani, Zhang, Dubois, Li, Guestrin, Liang, and Hashimoto}]{alpaca}
Rohan Taori, Ishaan Gulrajani, Tianyi Zhang, Yann Dubois, Xuechen Li, Carlos Guestrin, Percy Liang, and Tatsunori~B. Hashimoto. 2023.
\newblock Stanford alpaca: An instruction-following llama model.
\newblock \url{https://github.com/tatsu-lab/stanford_alpaca}.

\bibitem[{Touvron et~al.(2023)Touvron, Lavril, Izacard, Martinet, Lachaux, Lacroix, Rozi{\`e}re, Goyal, Hambro, Azhar et~al.}]{touvron2023llama}
Hugo Touvron, Thibaut Lavril, Gautier Izacard, Xavier Martinet, Marie-Anne Lachaux, Timoth{\'e}e Lacroix, Baptiste Rozi{\`e}re, Naman Goyal, Eric Hambro, Faisal Azhar, et~al. 2023.
\newblock \href {https://arxiv.org/abs/2302.13971} {Llama: Open and efficient foundation language models}.
\newblock \emph{ArXiv preprint}, abs/2302.13971.

\bibitem[{Voronov et~al.(2024)Voronov, Wolf, and Ryabinin}]{voronov2024mind}
Anton Voronov, Lena Wolf, and Max Ryabinin. 2024.
\newblock \href {https://arxiv.org/abs/2401.06766} {Mind your format: Towards consistent evaluation of in-context learning improvements}.
\newblock \emph{Preprint}, arXiv:2401.06766.

\bibitem[{Weber et~al.(2023)Weber, Bruni, and Hupkes}]{weber2023icl}
Lucas Weber, Elia Bruni, and Dieuwke Hupkes. 2023.
\newblock \href {https://arxiv.org/abs/2312.04945} {The icl consistency test}.
\newblock \emph{Preprint}, arXiv:2312.04945.

\bibitem[{Wei et~al.(2022)Wei, Bosma, Zhao, Guu, Yu, Lester, Du, Dai, and Le}]{weifinetuned}
Jason Wei, Maarten Bosma, Vincent~Y. Zhao, Kelvin Guu, Adams~Wei Yu, Brian Lester, Nan Du, Andrew~M. Dai, and Quoc~V. Le. 2022.
\newblock \href {https://openreview.net/forum?id=gEZrGCozdqR} {Finetuned language models are zero-shot learners}.
\newblock In \emph{The Tenth International Conference on Learning Representations, {ICLR} 2022, Virtual Event, April 25-29, 2022}. OpenReview.net.

\bibitem[{Wei et~al.(2023)Wei, Wang, Schuurmans, Bosma, Ichter, Xia, Chi, Le, and Zhou}]{wei2023chainofthought}
Jason Wei, Xuezhi Wang, Dale Schuurmans, Maarten Bosma, Brian Ichter, Fei Xia, Ed~Chi, Quoc Le, and Denny Zhou. 2023.
\newblock \href {https://arxiv.org/abs/2201.11903} {Chain-of-thought prompting elicits reasoning in large language models}.
\newblock \emph{Preprint}, arXiv:2201.11903.

\bibitem[{Ye et~al.(2024)Ye, Axmed, Pryzant, and Khani}]{ye2024prompt}
Qinyuan Ye, Maxamed Axmed, Reid Pryzant, and Fereshte Khani. 2024.
\newblock \href {https://arxiv.org/abs/2311.05661} {Prompt engineering a prompt engineer}.
\newblock \emph{Preprint}, arXiv:2311.05661.

\bibitem[{Zhao et~al.(2021)Zhao, Wallace, Feng, Klein, and Singh}]{pmlr-v139-zhao21c}
Zihao Zhao, Eric Wallace, Shi Feng, Dan Klein, and Sameer Singh. 2021.
\newblock \href {http://proceedings.mlr.press/v139/zhao21c.html} {Calibrate before use: Improving few-shot performance of language models}.
\newblock In \emph{Proceedings of the 38th International Conference on Machine Learning, {ICML} 2021, 18-24 July 2021, Virtual Event}, volume 139 of \emph{Proceedings of Machine Learning Research}, pages 12697--12706. {PMLR}.

\bibitem[{Zhou et~al.(2023)Zhou, Muresanu, Han, Paster, Pitis, Chan, and Ba}]{zhou2023large}
Yongchao Zhou, Andrei~Ioan Muresanu, Ziwen Han, Keiran Paster, Silviu Pitis, Harris Chan, and Jimmy Ba. 2023.
\newblock \href {https://arxiv.org/abs/2211.01910} {Large language models are human-level prompt engineers}.
\newblock \emph{Preprint}, arXiv:2211.01910.

\bibitem[{Zhu et~al.(2023)Zhu, Wang, Zhou, Wang, Chen, Wang, Yang, Ye, Zhang, Gong, and Xie}]{zhu2023promptbench}
Kaijie Zhu, Jindong Wang, Jiaheng Zhou, Zichen Wang, Hao Chen, Yidong Wang, Linyi Yang, Wei Ye, Yue Zhang, Neil~Zhenqiang Gong, and Xing Xie. 2023.
\newblock \href {https://arxiv.org/abs/2306.04528} {Promptbench: Towards evaluating the robustness of large language models on adversarial prompts}.
\newblock \emph{Preprint}, arXiv:2306.04528.

\end{thebibliography}
\newpage
\appendix
\section{Code and Data}
\label{app:code_and_data}
Our code that we used for generating the variants of prompts and computing the sensitivity metric is open-sourced at \href{https://github.com/kowndinya-renduchintala/POSIX}{https://github.com/kowndinya-renduchintala/POSIX}. The code development utilized open-source tools, primarily relying on the HuggingFace library for inference, with PyTorch as the underlying framework. Both PyTorch and HuggingFace are licensed under permissive licenses, with PyTorch under the BSD license and HuggingFace under the Apache 2.0 license.

\section{Prompt Templates}\label{appendix:templates}
The original template used for experiments on MMLU is: \begin{center}
    \texttt{Q:\{\}\textbackslash n(A)\{\} (B)\{\} (C)\{\} (D)\{\}\textbackslash nA: }
\end{center}
The other \(20\) prompt templates used for the experiments with template variations on MMLU are listed in Table \ref{tab:mmlu_templates}.\\

\noindent For experiments on the Alpaca dataset which contains open-ended questions, the default template used is:\hspace{4mm} \texttt{Q:\{\}\textbackslash nA: }\hspace{0.5mm}. The other \(20\) prompt templates used for the experiments with template variations on Alpaca are listed in Table \ref{tab:alpaca_templates}.

\begin{table}[!h]
\begin{center}
\scalebox{0.7}{
\begin{tabular}{ |c|c|}
 \hline
 {\bf Seed}&{\bf Prompt Template} \\ \hline
 0 & \texttt{q:\{\}\textbackslash n(A)\{\} (B)\{\} (C)\{\} (D)\{\}\textbackslash na: } \\ \hline

1 & \texttt{Q::\{\}\textbackslash n(A)\{\} (B)\{\} (C)\{\} (D)\{\}\textbackslash nA:: } \\ \hline

2 & \texttt{Q:  \{\}\textbackslash n(A)\{\} (B)\{\} (C)\{\} (D)\{\}\textbackslash nA:  } \\ \hline

3 & \texttt{q:: \{\}\textbackslash n(A)\{\} (B)\{\} (C)\{\} (D)\{\}\textbackslash na:: } \\ \hline

4 & \texttt{Q::: \{\}\textbackslash n(A)\{\} (B)\{\} (C)\{\} (D)\{\}\textbackslash nA::: }\\ \hline

5 & \texttt{Q: \{\} || (A)\{\} (B)\{\} (C)\{\} (D)\{\} || A: }\\ \hline

6& \texttt{q:::\{\}\textbackslash n(A)\{\} (B)\{\} (C)\{\} (D)\{\}\textbackslash na::: }\\ \hline

7& \texttt{Q: \{\}\textbackslash n(A)\{\} (B)\{\} (C)\{\} (D)\{\}\textbackslash nAnswer: }\\ \hline

8& \texttt{QUESTION:\{\}\textbackslash n(A)\{\} (B)\{\} (C)\{\} (D)\{\}\textbackslash nA: }\\ \hline

9& \texttt{Question:\{\}\textbackslash n(A)\{\} (B)\{\} (C)\{\} (D)\{\}\textbackslash nAnswer: }\\ \hline
    
10& \texttt{QUESTION:\{\}\textbackslash n(A)\{\} (B)\{\} (C)\{\} (D)\{\}\textbackslash nANSWER: }\\ \hline
    
11& \texttt{Question: \{\}\textbackslash n(A)\{\} (B)\{\} (C)\{\} (D)\{\}\textbackslash nAnswer: }\\ \hline
    
12& \texttt{Question::: \{\}\textbackslash n(A)\{\} (B)\{\} (C)\{\} (D)\{\}\textbackslash nAnswer::: }\\ \hline
    
13& \texttt{QUESTION: \{\}\textbackslash n(A)\{\} (B)\{\} (C)\{\} (D)\{\}\textbackslash nAnswer: }\\ \hline
    
14& \texttt{Question - \{\}\textbackslash n(A)\{\} (B)\{\} (C)\{\} (D)\{\}\textbackslash nAnswer - }\\ \hline
    
15& \texttt{question::\{\}\textbackslash n(A)\{\} (B)\{\} (C)\{\} (D)\{\}\textbackslash nanswer:: }\\ \hline
    
16& \texttt{question: \{\}\textbackslash n(A)\{\} (B)\{\} (C)\{\} (D)\{\}\textbackslash nanswer: }\\ \hline
    
17& \texttt{Question: \{\} || (A)\{\} (B)\{\} (C)\{\} (D)\{\} || Answer: }\\ \hline
    
18 & \texttt{QUESTION\textbackslash t\{\}\textbackslash n(A)\{\} (B)\{\} (C)\{\} (D)\{\}\textbackslash nANSWER\textbackslash t}\\ \hline
    
19 & \texttt{Question: \{\} , (A)\{\} (B)\{\} (C)\{\} (D)\{\} , Answer: } \\

\hline
\end{tabular}
}
\caption{Prompt Templates used for MMLU.}
\vspace{-3mm}
\label{tab:mmlu_templates}
\end{center}
\end{table}

\begin{table}[!h]
\begin{center}
\scalebox{0.75}{
\begin{tabular}{ |c|c|}
 \hline
 {\bf Seed}&{\bf Prompt Template} \\ \hline
 0 & \texttt{q:\{\}\textbackslash na: } \\ \hline

1 & \texttt{Q::\{\}\textbackslash nA:: } \\ \hline

2 & \texttt{Q:  \{\}\textbackslash nA:  } \\ \hline

3 & \texttt{q:: \{\}\textbackslash na:: } \\ \hline

4 & \texttt{Q::: \{\}\textbackslash nA::: }\\ \hline

5 & \texttt{Q: \{\} || A: }\\ \hline

6& \texttt{q:::\{\}\textbackslash na::: }\\ \hline

7& \texttt{Q: \{\}\textbackslash nAnswer: }\\ \hline

8& \texttt{QUESTION:\{\}\textbackslash nA: }\\ \hline

9& \texttt{Question:\{\}\textbackslash nAnswer: }\\ \hline
    
10& \texttt{QUESTION:\{\}\textbackslash nANSWER: }\\ \hline
    
11& \texttt{Question: \{\}\textbackslash nAnswer: }\\ \hline
    
12& \texttt{Question::: \{\}\textbackslash nAnswer::: }\\ \hline
    
13& \texttt{QUESTION: \{\}\textbackslash nAnswer: }\\ \hline
    
14& \texttt{Question - \{\}\textbackslash nAnswer - }\\ \hline
    
15& \texttt{question::\{\}\textbackslash nanswer:: }\\ \hline
    
16& \texttt{question: \{\}\textbackslash nanswer: }\\ \hline
    
17& \texttt{Question: \{\} || Answer: }\\ \hline
    
18 & \texttt{QUESTION\textbackslash t\{\}\textbackslash nANSWER\textbackslash t}\\ \hline
    
19 & \texttt{Question: \{\} , Answer: } \\

\hline
\end{tabular}
}
\caption{Prompt Templates used for Alpaca.}
\vspace{-3mm}
\label{tab:alpaca_templates}
\end{center}
\end{table}
Also, for experiments on MMLU, we prepend the following instruction before the template: \\

    \texttt{The following are multiple choice questions (with answers) about \{subject\}.\textbackslash n\textbackslash n}

The \texttt{\{subject\}} is filled with the corresponding topic name from MMLU. 


\section{Sensitivity of OLMo-1B}\label{appendix:olmo1b}
Table \ref{tab:olmo1b} reports the \metric\ values of OLMo-1B model for different variation types on MMLU and Alpaca.

\begin{table}[h!]

    \centering
    \scalebox{0.8}{
        \begin{tabular}{lcc}
            \toprule
            Variation Type & MMLU-ZeroShot & Alpaca-ZeroShot\\
            \midrule
            
            Spelling Errors & $0.089_{\pm 0.099}$ & $0.257_{\pm 0.229}$  \\
            
            Prompt Templates & $0.896_{\pm 0.205}$ & $0.355_{\pm 0.069}$  \\
            
            Paraphrases & $0.102_{\pm 0.082}$ & $0.185_{\pm 0.128}$ \\
            
            Mixture & $0.602_{\pm 0.17}$ & $0.358_{\pm 0.139}$ \\
            \hline
        \end{tabular}
    }
    \caption{Sensitivity of OLMo-1B.}
    \label{tab:olmo1b}
\end{table}

\section{Sensitivity of Llama-2-13B}\label{appendix:llama13b}
Table \ref{tab:llama-2-13b-MMLU} reports the \metric\ values of Llama-2-13B models (base and chat variants) for different variation types on MMLU and Table \ref{tab:llama-2-13b-Alpaca} reports it for Alpaca dataset.

\begin{table}[h!]

    \centering
    \scalebox{0.8}{
        \begin{tabular}{lcc}
            \toprule
            Variation Type & Llama-2-13B & Llama-2-Chat-13B\\
            \midrule
            
            Spelling Errors & $0.057_{\pm 0.073}$ & $0.072_{\pm 0.086}$ \\
            
            Prompt Templates & $1.16_{\pm 0.518}$ & $0.858_{\pm 0.324}$ \\
            
            Paraphrases & $0.093_{\pm 0.106}$ & $0.095_{\pm 0.102}$ \\
            
            Mixture & $0.677_{\pm 0.349}$ & $0.438_{\pm 0.248}$ \\
            \hline
        \end{tabular}
    }
    \caption{Sensitivity of Llama-2-13B (MMLU-ZeroShot)}
    \label{tab:llama-2-13b-MMLU}
\end{table}

\begin{table}[h!]

    \centering
    \scalebox{0.8}{
        \begin{tabular}{lcc}
            \toprule
            Variation Type & Llama-2-13B & Llama-2-Chat-13B\\
            \midrule
            
            Spelling Errors & $0.141_{\pm 0.126}$ & $0.222_{\pm 0.172}$ \\
            
            Prompt Templates & $0.177_{\pm 0.109}$ & $0.134_{\pm 0.135}$ \\
            
            Paraphrases & $0.225_{\pm 0.173}$ & $0.592_{\pm 0.3}$ \\
            
            Mixture & $0.249_{\pm 0.148}$ & $0.461_{\pm 0.213}$ \\
            \hline
        \end{tabular}
    }
    \caption{Sensitivity of Llama-2-13B (Alpaca-ZeroShot)}
    \label{tab:llama-2-13b-Alpaca}
\end{table}

\section{Sensitivity on the BBH dataset}\label{appendix:bbh}
\begin{figure}
    \centering
\includegraphics[width=\linewidth]{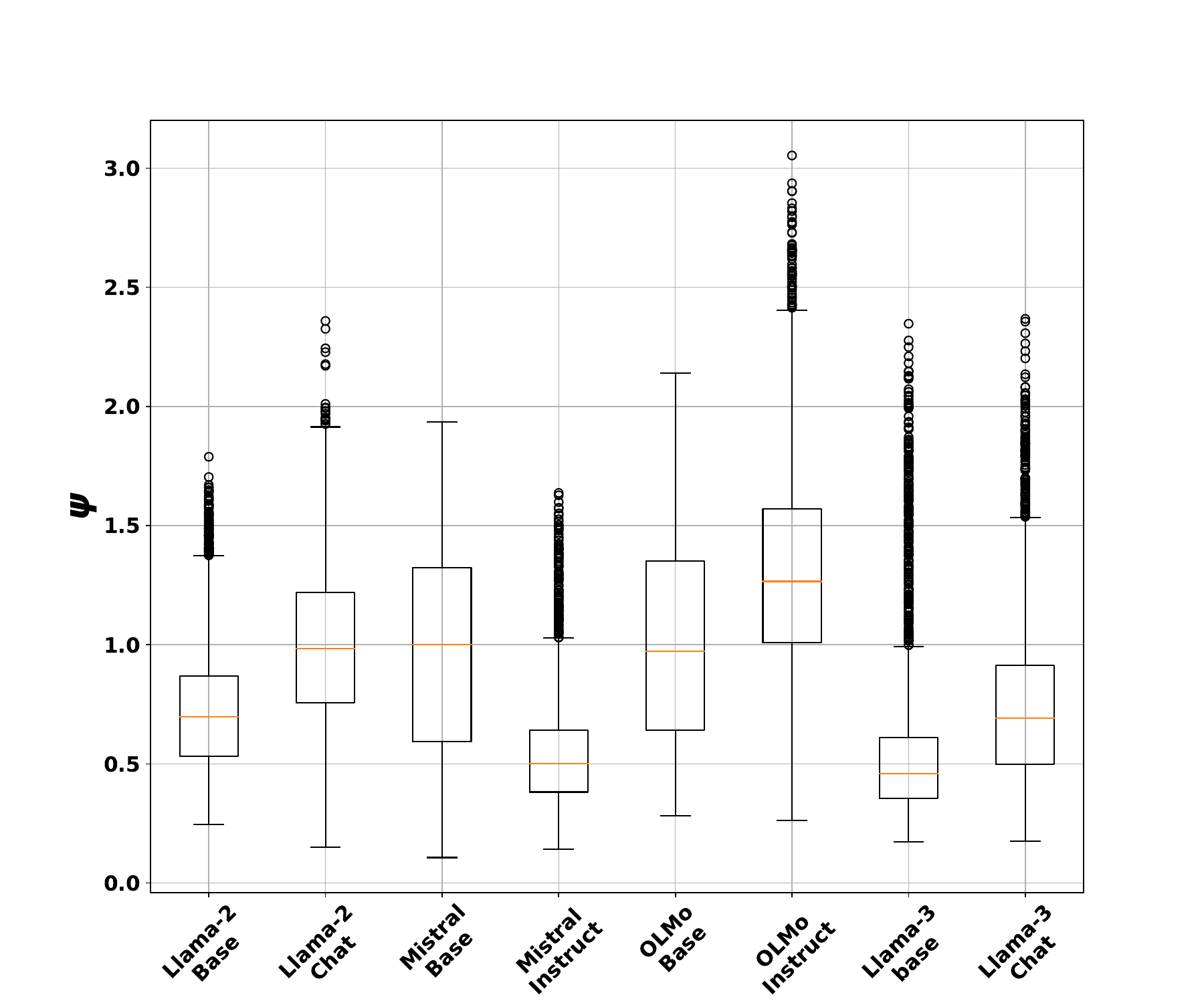}
    \caption{Box plots depicting distribution of $\psi_{\mathcal{M}, \mathbf{X}}$ for various 7B models on Big Bench Hard (BBH) dataset}
    \label{fig:bbh_box_plots}
\end{figure}

\begin{figure}
    \centering
\includegraphics[width=\linewidth]{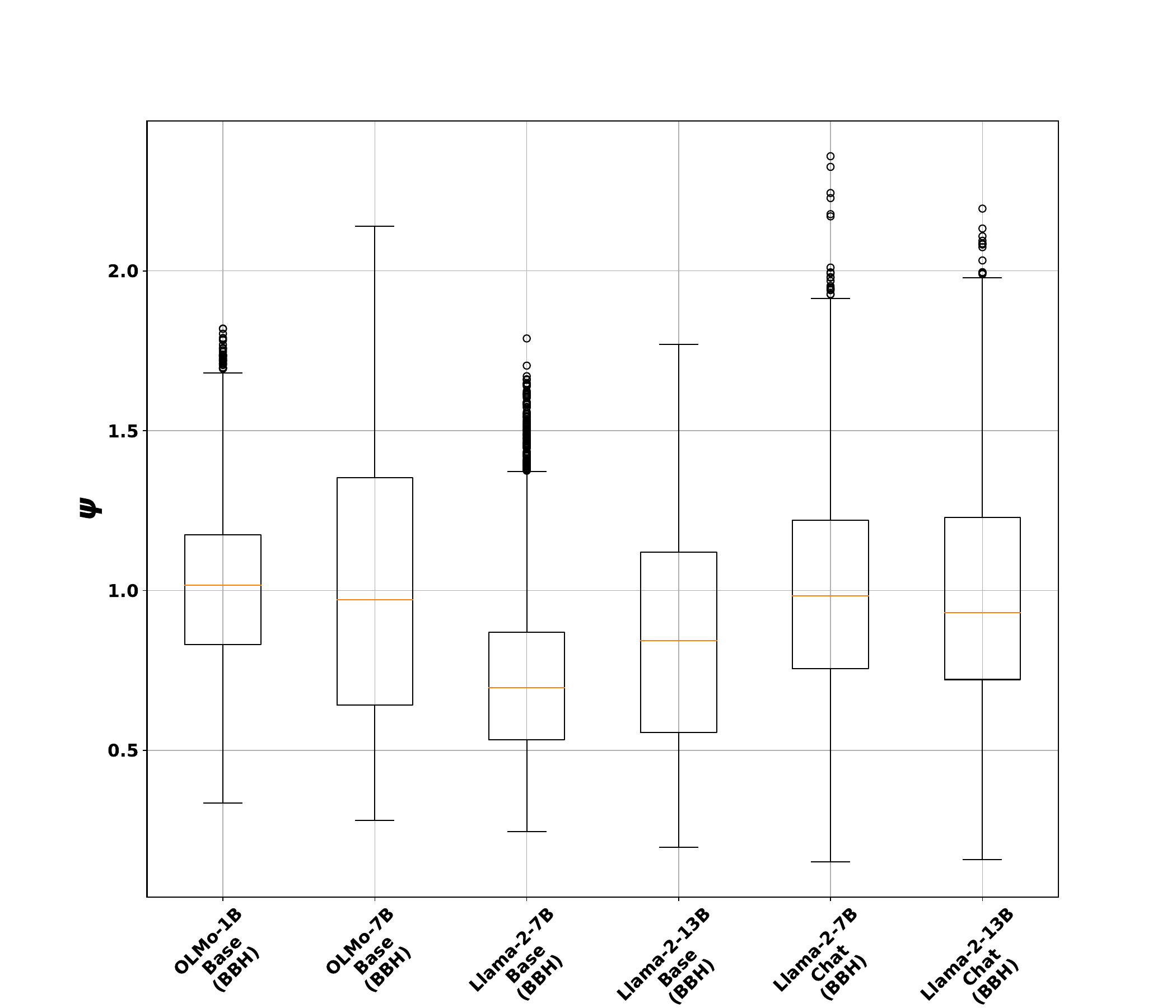}
    \caption{Box plots depicting distribution of $\psi_{\mathcal{M}, \mathbf{X}}$ when varying model scale, for OLMo (1B and 7B) and Llama-2 (7B and 13B) -- both base and chat variants.}
    \label{fig:bbh_model_scale_figure}
\end{figure}

Table \ref{tab:bbh} reports the \metric\ values on the Big-Bench Hard (BBH) dataset. Additionally, Figure~\ref{fig:bbh_box_plots} depicts prompt sensitivity of various models on BBH dataset in the form of box plots and Figure~\ref{fig:bbh_model_scale_figure} depicts the effect of varying model scale on prompt sensitivity for BBH dataset. Please note that we only consider one kind of variation - prompt template - for this dataset. This is because in the case of BBH, many tasks like boolean\_expressions, date understanding, geometric\_shapes or dyck\_languages, even minor spelling errors are not intent-preserving and paraphrasing would not be possible in many cases such as numerical expressions. For this dataset, we sample 2700 samples randomly from 23 tasks. 

\begin{table}[h!]

    \centering
    \scalebox{0.8}{
        \begin{tabular}{lc}
            \toprule
            Model & BBH-ZeroShot \\
            \midrule
            
            Llama-2-7b & $0.729_{\pm 0.295}$ \\
            Llama-2-7b-chat & $0.989_{\pm 0.351}$\\
            Llama-3-8b & $0.58_{\pm 0.385}$ \\
            Llama-3-8b-chat & $0.745_{\pm 0.35}$\\
            Mistral-7B & $0.966_{\pm 0.447}$ \\
            Mistral-7B-Instruct & $0.542_{\pm 0.239}$ \\
            OLMo-7B-Base & $1.029_{\pm 0.427}$ \\
            OLMo-7B-Instruct & $1.304_{\pm 0.47}$ \\
            OLMo-1B & $1.021_{\pm 0.292}$ \\
            Llama-2-13b & $0.851_{\pm 0.357}$ \\
            Llama-2-13b-chat & $0.972_{\pm 0.388}$ \\
            \hline
        \end{tabular}
    }
    \caption{Sensitivity of various models on the BBH dataset}
    \label{tab:bbh}
\end{table}

\section{Examples of generated paraphrases}
\label{appendix:paraphrase_examples}
Please note that we open source all the data at \href{https://github.com/kowndinya-renduchintala/POSIX}{https://github.com/kowndinya-renduchintala/POSIX}. Although, following are some examples of generated paraphrases for few open-ended questions from alpaca data:
\begin{itemize}[nosep]
    \item \textbf{Original Question:} How much do you know about Buddhism?
    \begin{itemize}[nosep]
        \item What is your awareness of Buddhism?
        \item What is your level of expertise on Buddhism?
        \item Are you familiar with the principles of Buddhism?
        \item What is your level of familiarity with Buddhism?
        \item Can you share your knowledge of Buddhism with me?
    \end{itemize}
    \item \textbf{Original Question:} Explain the concept of cognitive biases.
    \begin{itemize}[nosep]
        \item Interpret the idea of cognitive biases
        \item Expound on the concept of cognitive biases
        \item Elaborate on the concept of cognitive biases
        \item Explicate the concept of cognitive biases
        \item Spell out the notion of cognitive biases
    \end{itemize}
    \item \textbf{Original Question:} Describe the best way to store fresh berries.
    \begin{itemize}[nosep]
        \item Provide instructions on how to store fresh berries for maximum freshness.
        \item Offer advice on how to best store fresh berries.
        \item Elaborate on the best way to store fresh berries to maintain their freshness.
        \item Detail the optimal way to keep fresh berries fresh for longer.
        \item Elaborate on the proper way to store a bunch of fresh berries.
    \end{itemize}
\end{itemize}

\section{Efficacy of \metric\ for Open-ended Generation}
\label{app:efficacy_figure_alpaca}
\begin{figure*}[thb!]
    \centering
    \begin{subfigure}[b]{3.1in}
        \centering
        \includegraphics[width=3.48in]{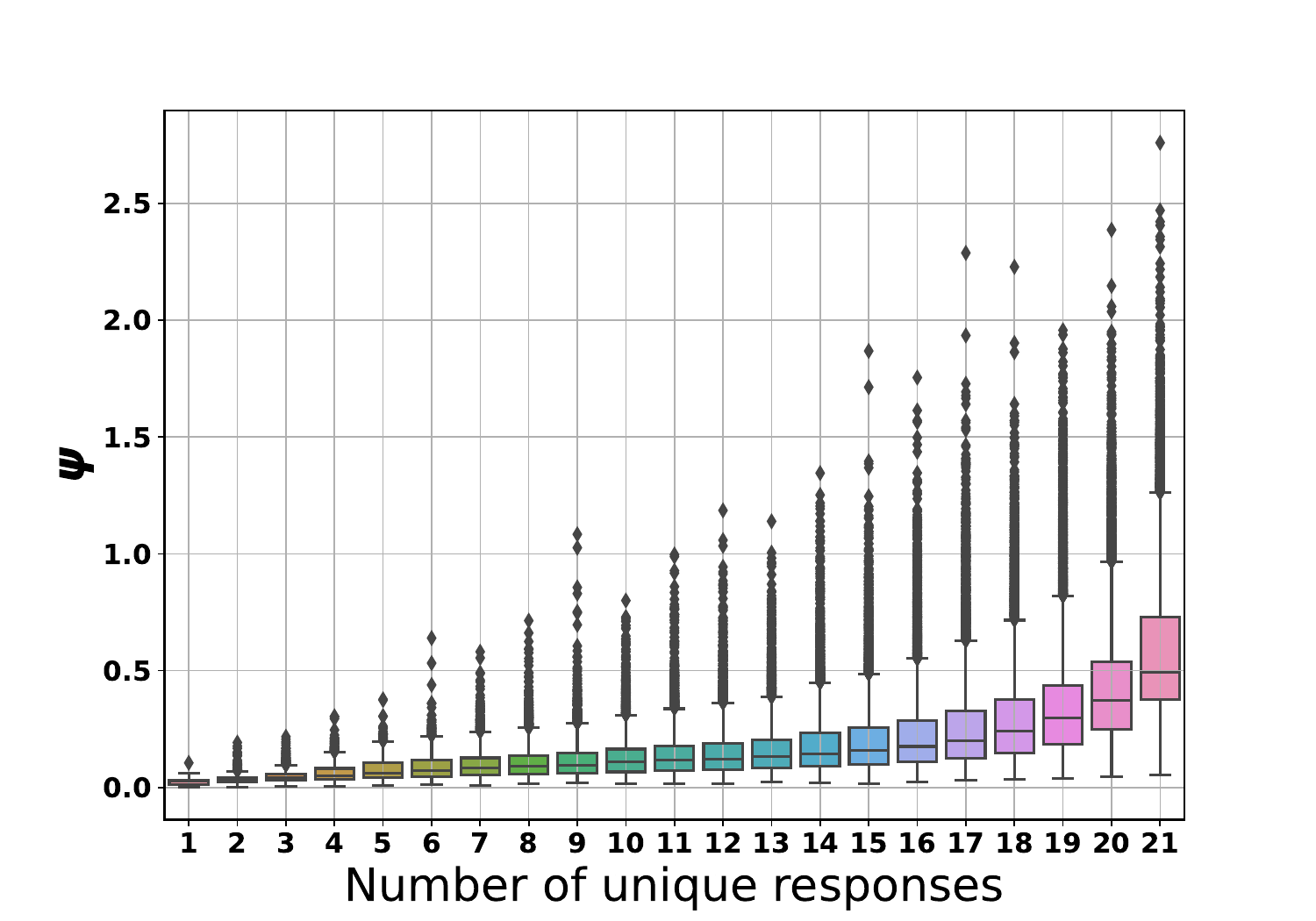}
        \caption{}
    \end{subfigure}
    \begin{subfigure}[b]{3.1in}
        \includegraphics[width=3.48in]{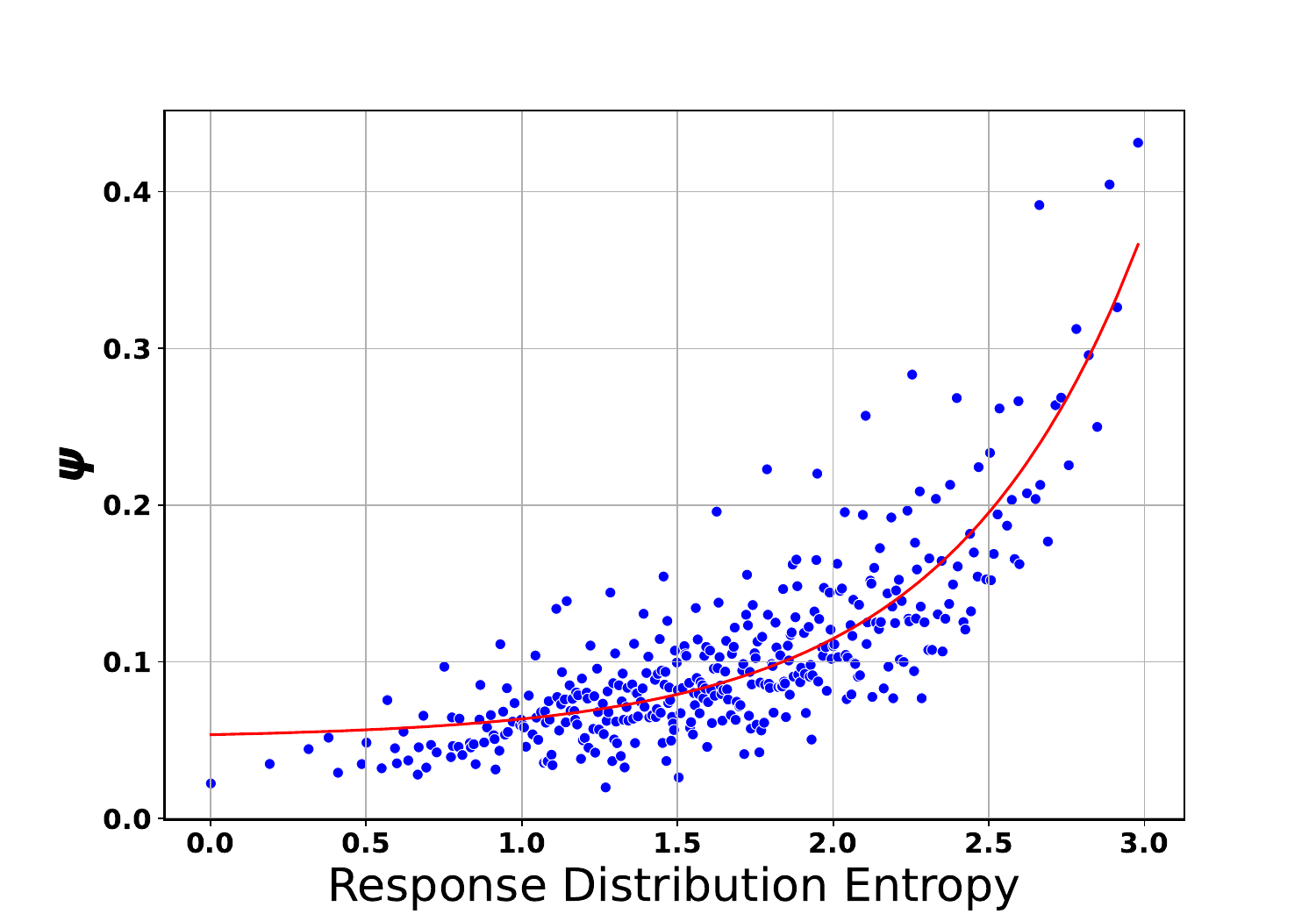}
        \caption{}
    \end{subfigure}
    \begin{subfigure}[b]{3.1in}
        \includegraphics[width=3.48in]{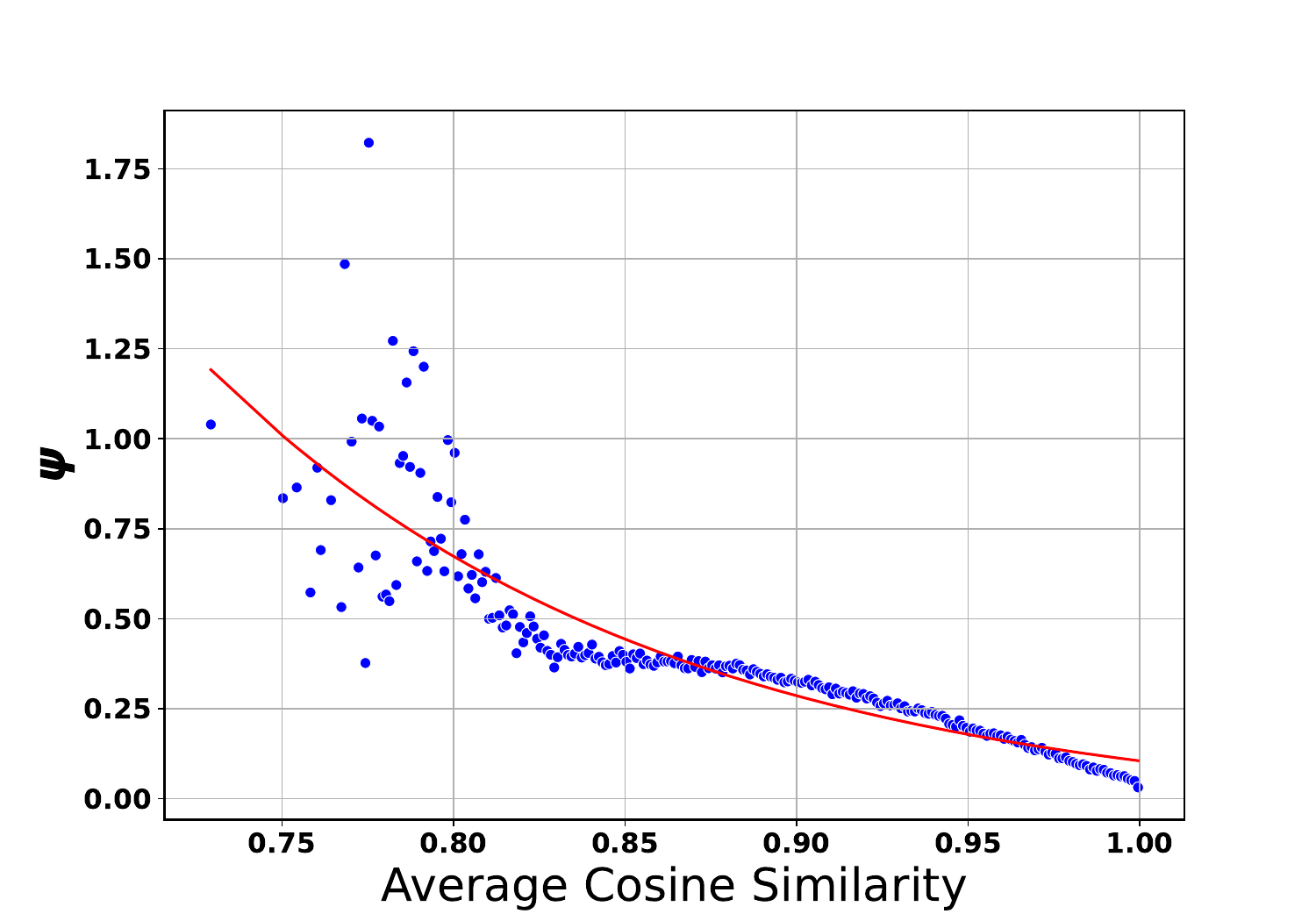}
        \caption{}
    \end{subfigure}
    \begin{subfigure}[b]{0.46\textwidth}
        \includegraphics[width=3.48in]
        {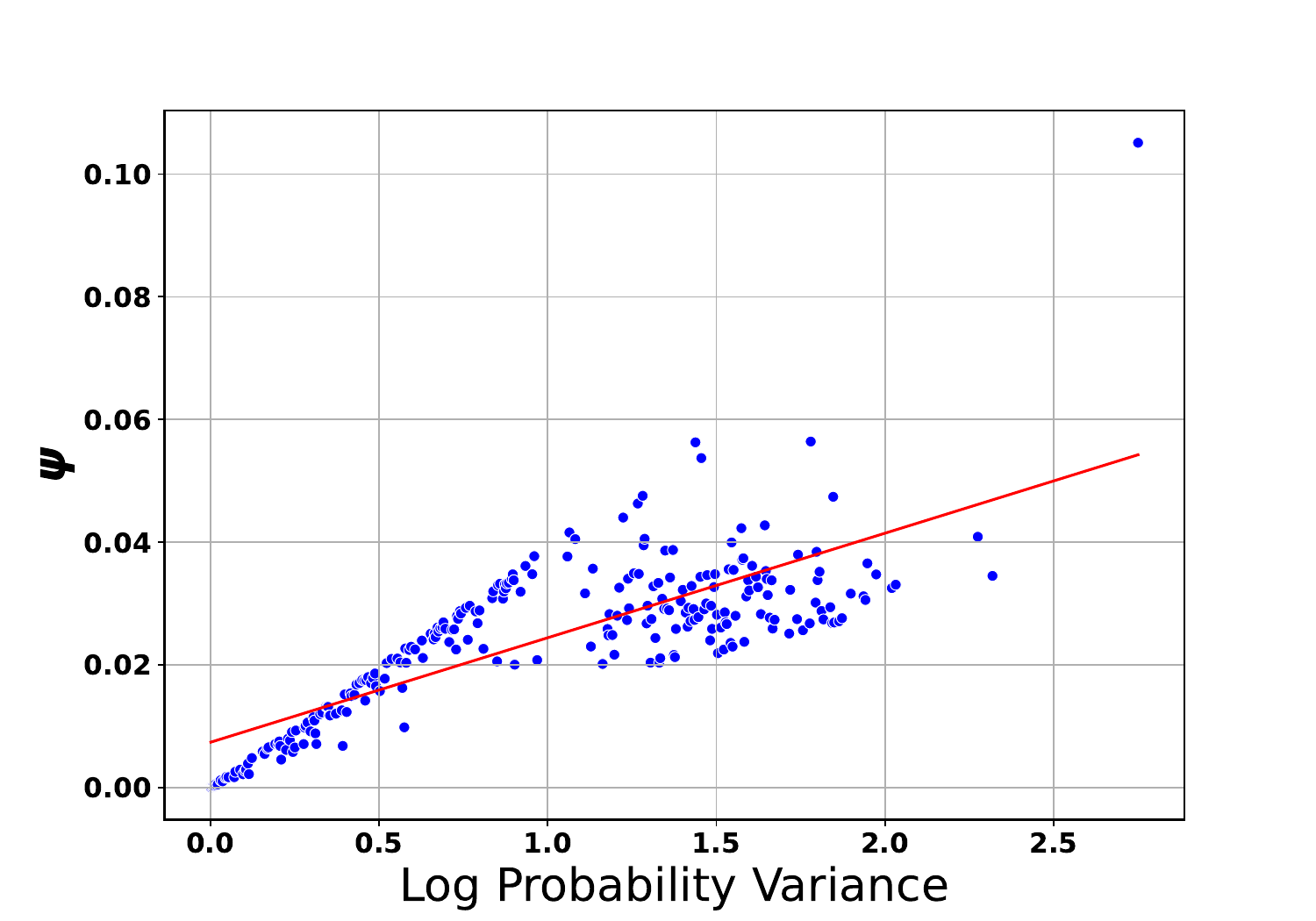}
        \caption{}
    \end{subfigure}
    \caption{Correlation plots of $\psi$ with each of the four factors described in Section~\ref{sec:what_does_posix_capture} in the case of Alpaca: (a) Response Diversity; (b) Response Distribution Entropy; (c) Semantic Coherence; (d) Variance in Confidence. }
    \label{fig:efficacy_alpaca}
\end{figure*}
Figure \ref{fig:efficacy_alpaca} shows the correlation of \metric\ with the four factors listed in Section \ref{sec:introduction} for a combination of all types of prompt variations in Alpaca, depicting the effectiveness of \metric\ in successfully capturing the nuances of prompt sensitivity.

\section{Distribution of \metric\ Values for All Variation Types}\label{appendix:box_plots}
\begin{figure}[h!]
    \centering
        \includegraphics[width=0.9\linewidth]{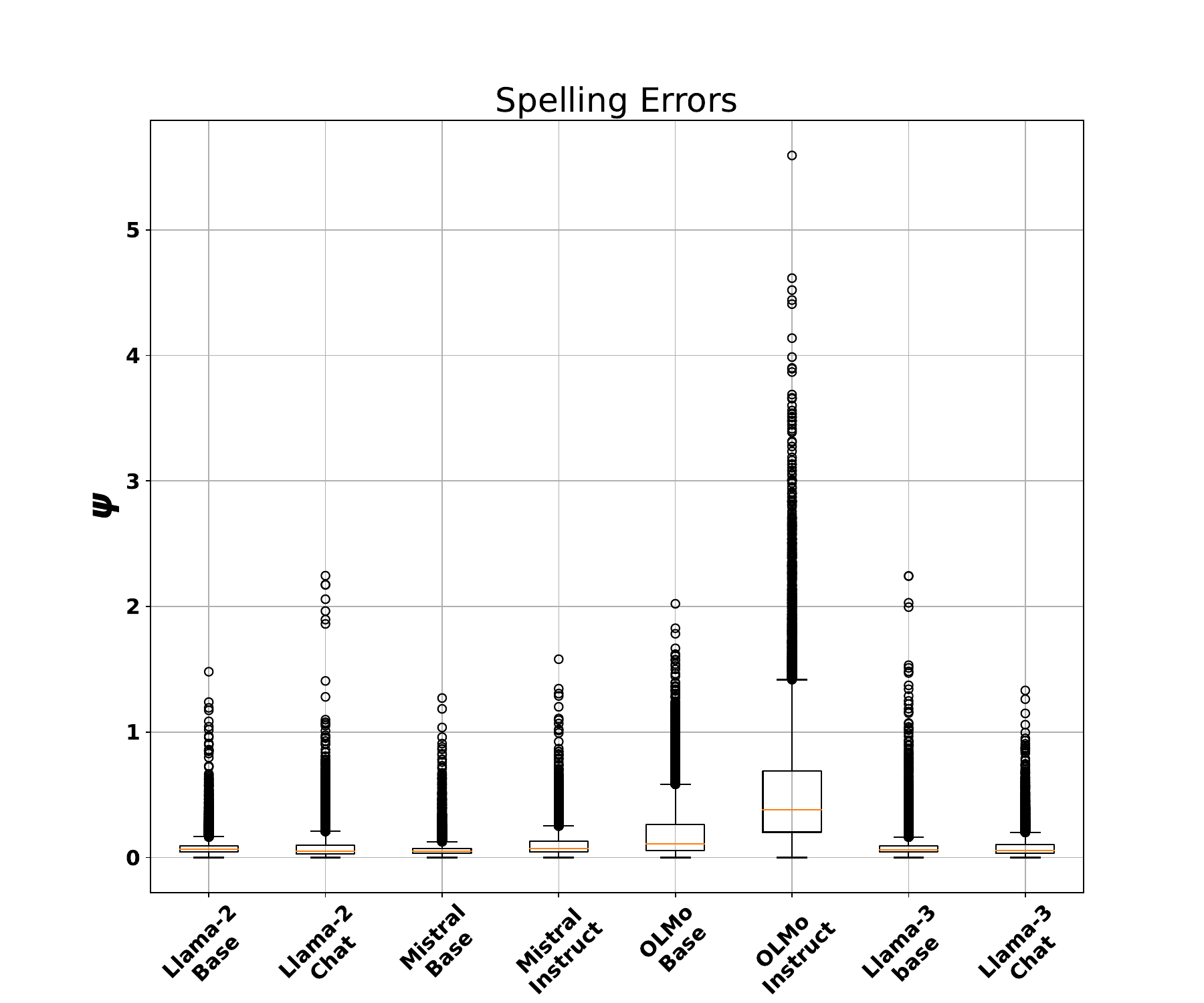}
        \includegraphics[width=0.9\linewidth]{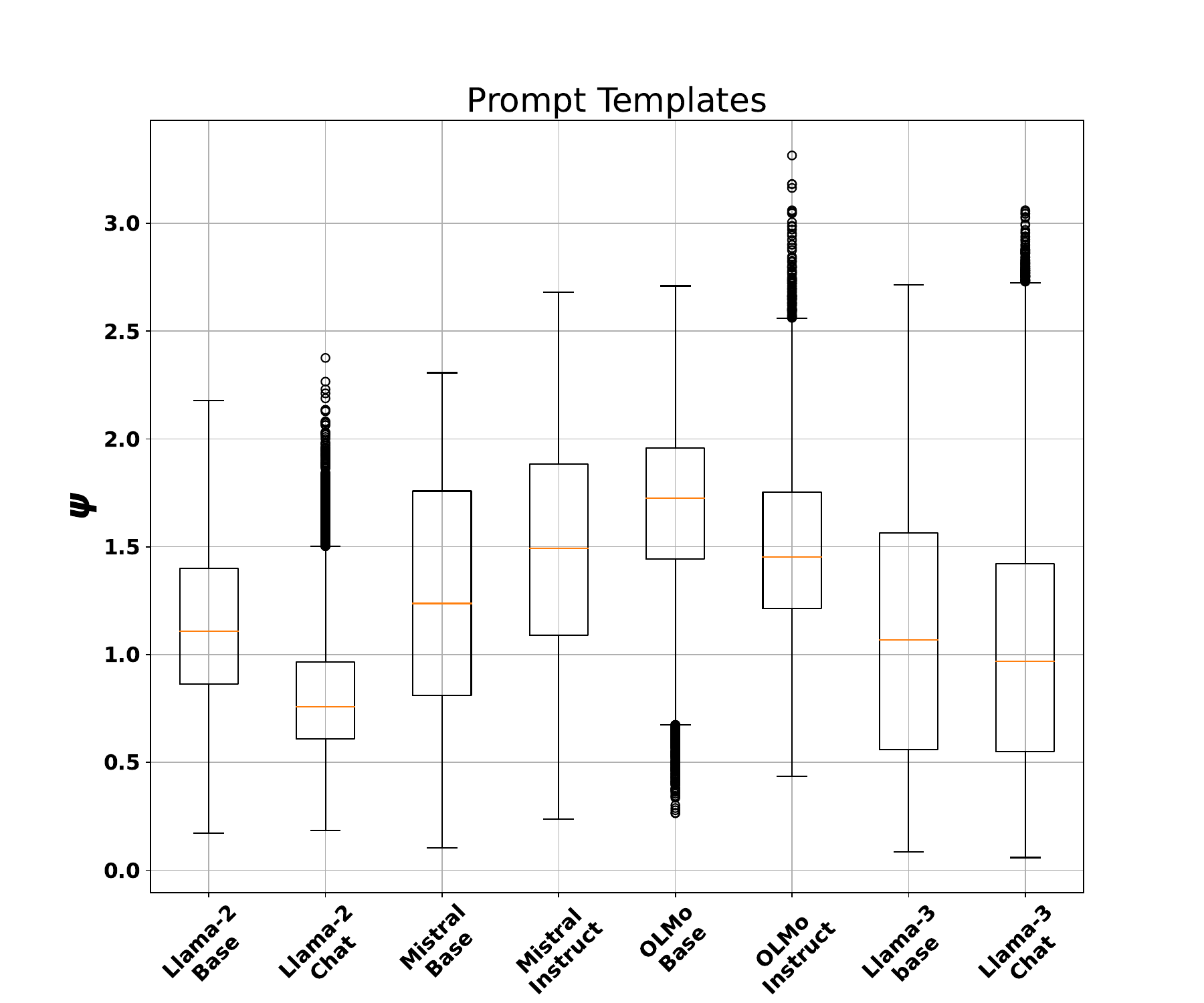}
        \includegraphics[width=0.9\linewidth]{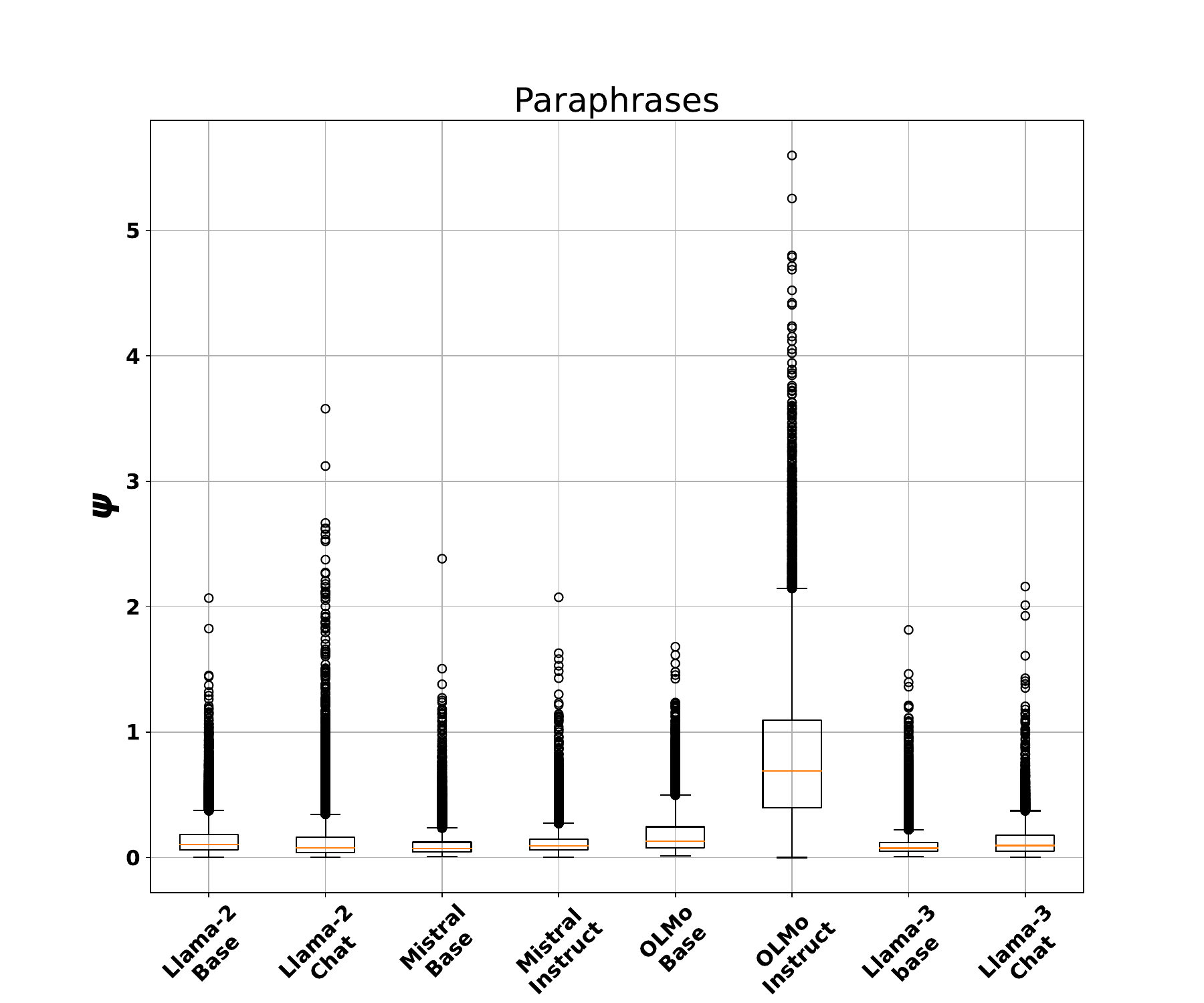}
        \includegraphics[width=0.9\linewidth]{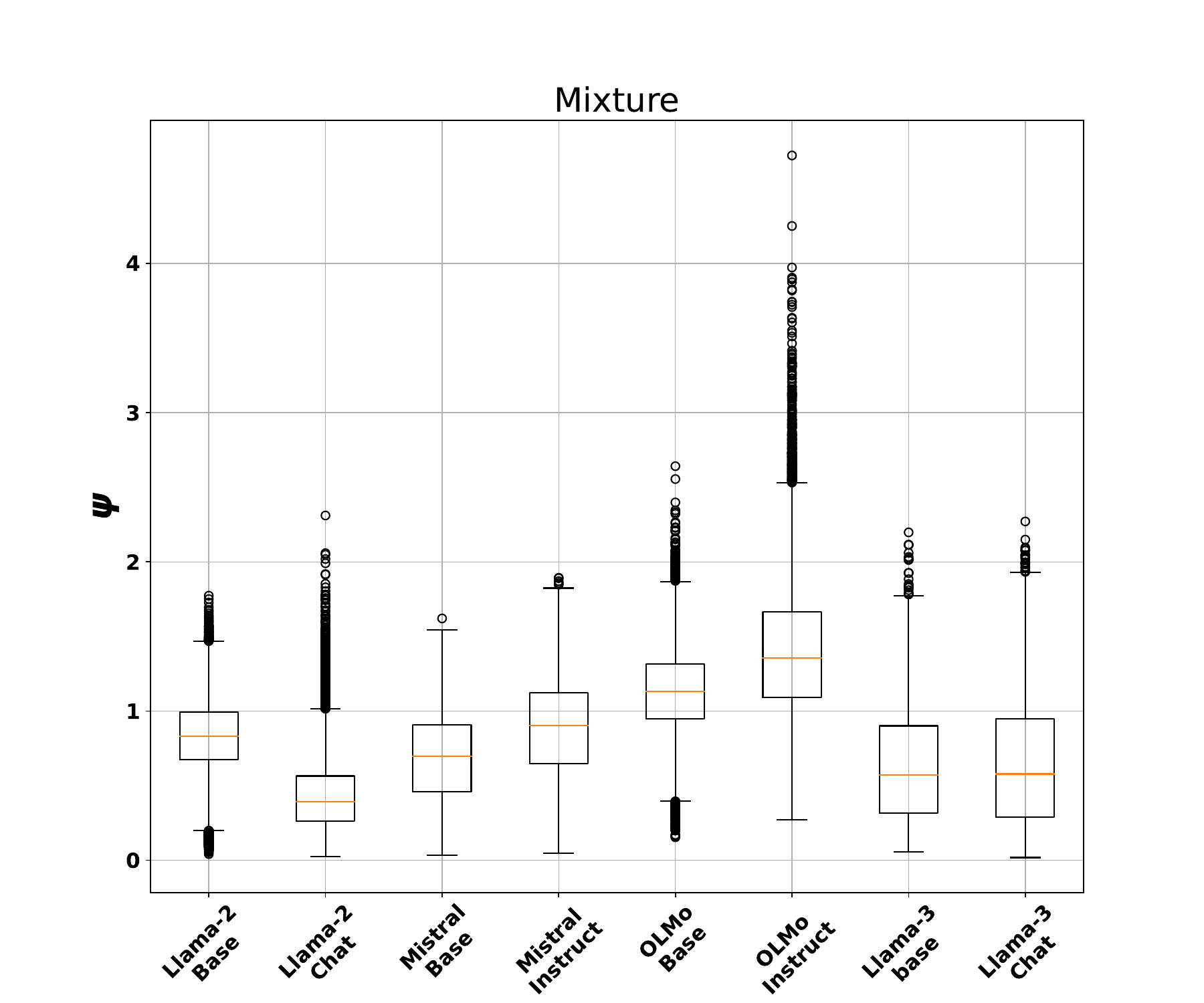}
    
    \caption{Box Plots depicting variation of $\psi$ for different prompt variations in case of MMLU.}
    \label{fig:box_plots_all_mmlu}
\end{figure}

\begin{figure}[h!]
    \centering
    \includegraphics[width=0.9\linewidth]{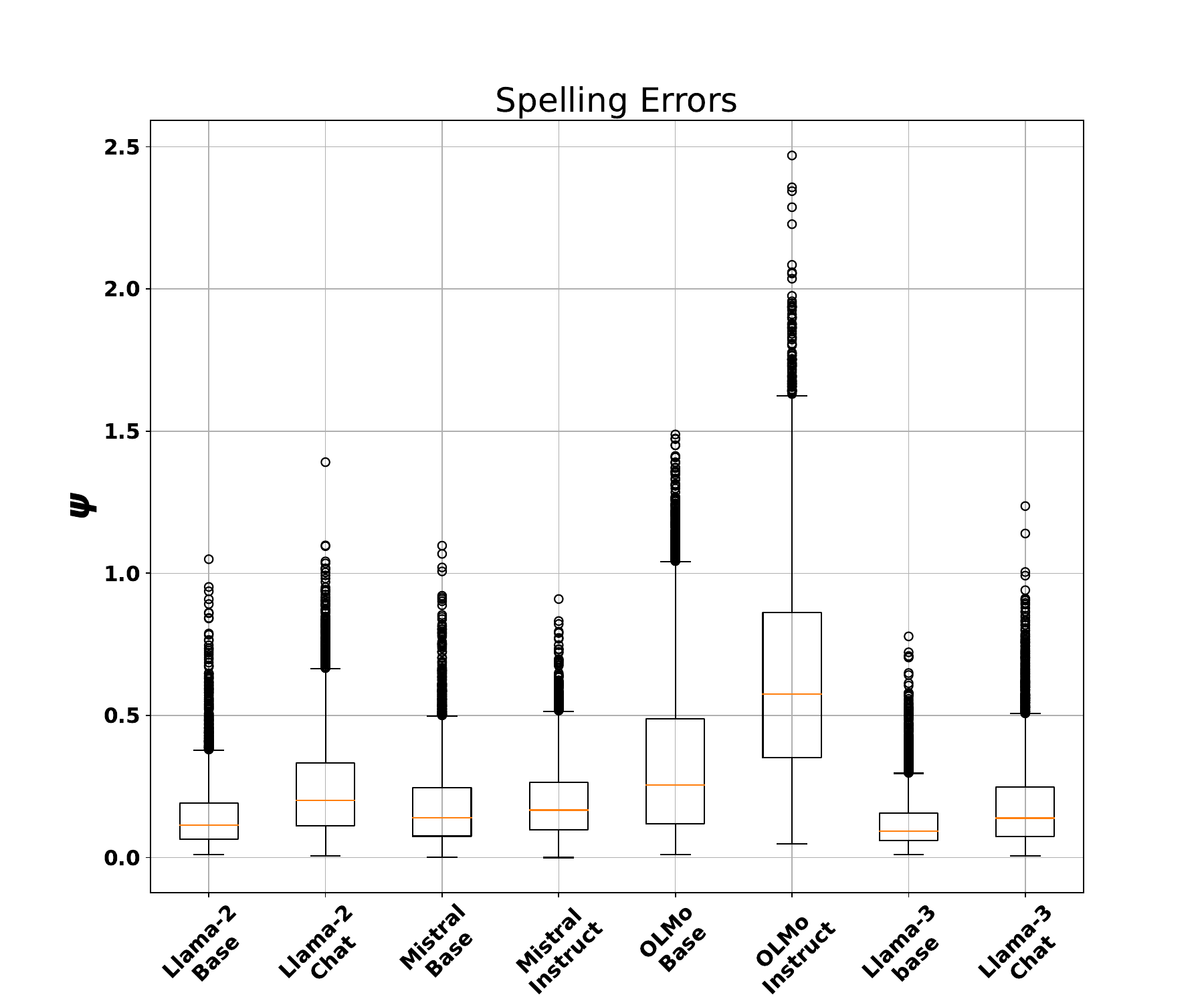}
    \includegraphics[width=0.9\linewidth]{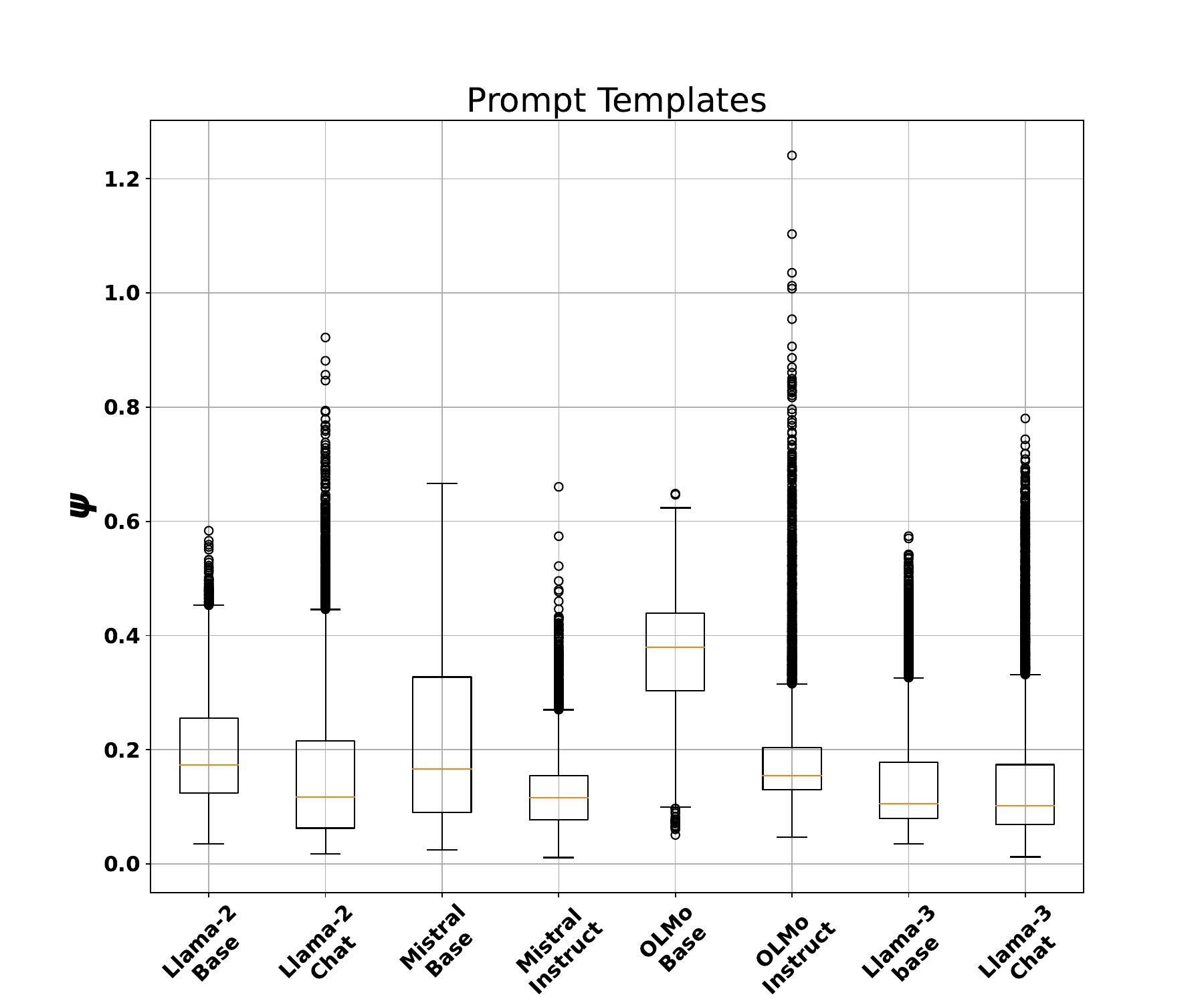}
    \includegraphics[width=0.9\linewidth]{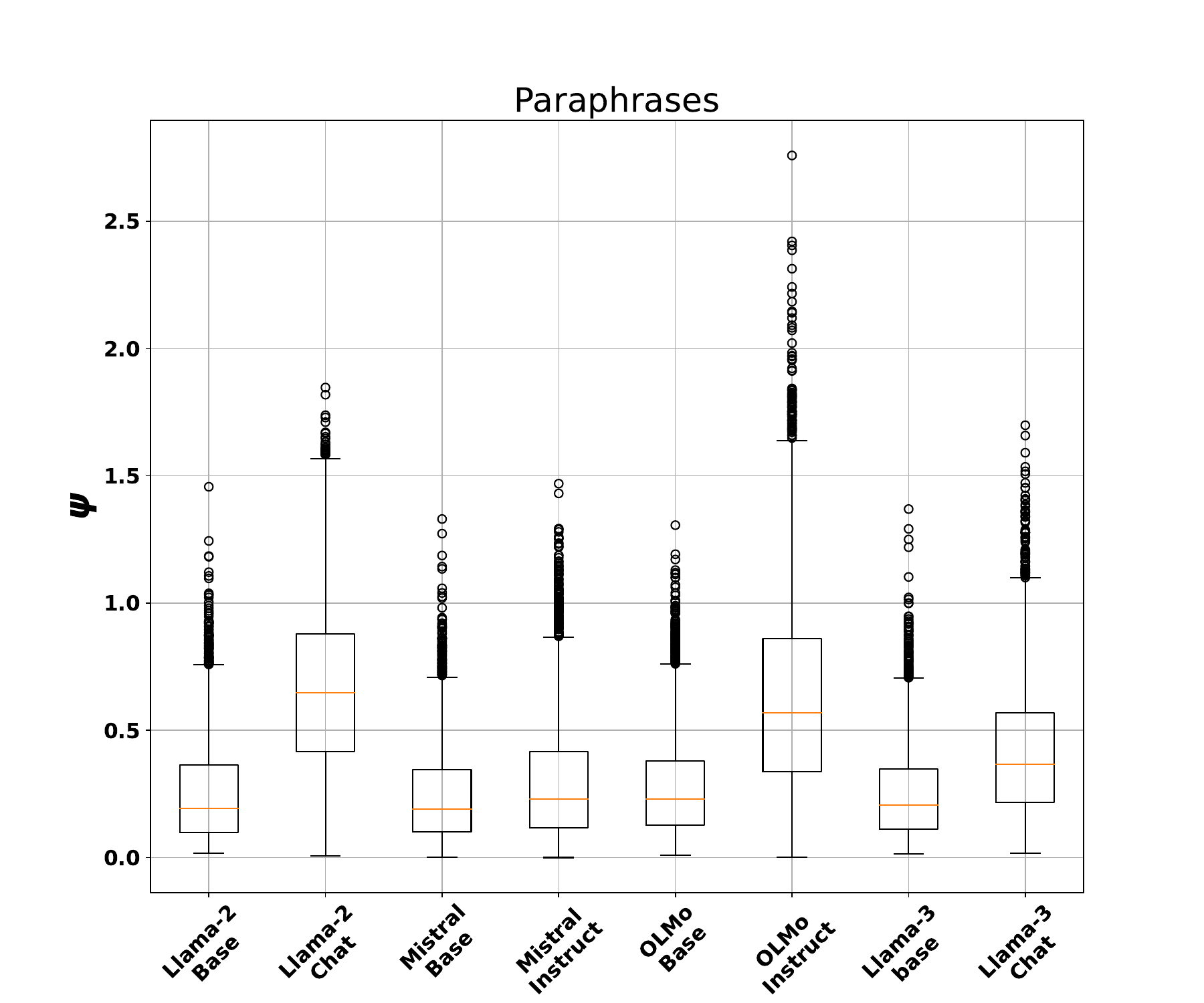}
    \includegraphics[width=0.9\linewidth]{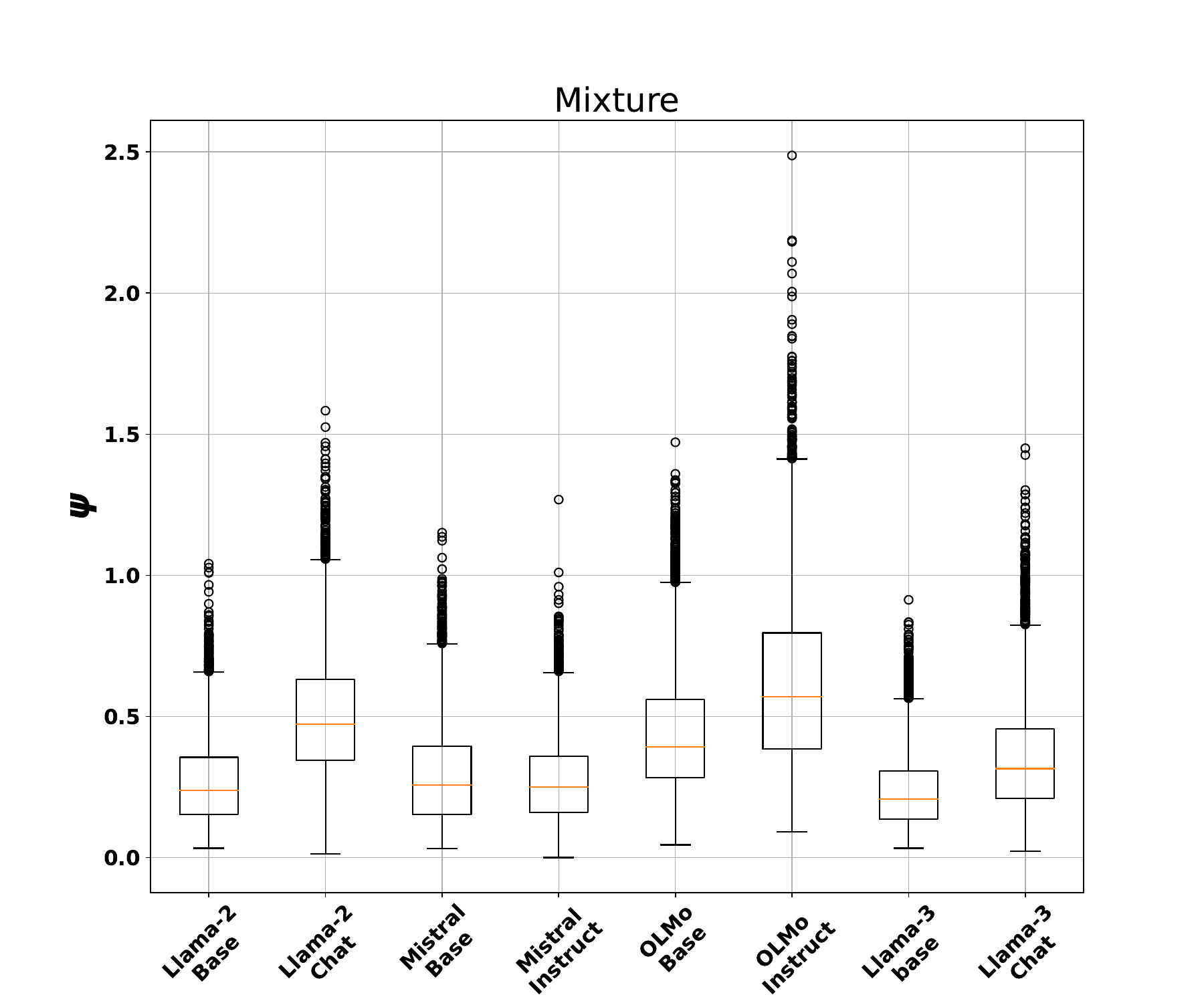}
    \caption{Box Plots depicting variation of $\psi$ for different prompt variations in case of Alpaca.}
    \label{fig:box_plots_all_alpaca}
\end{figure}

Figure \ref{fig:box_plots_all_mmlu} and Figure \ref{fig:box_plots_all_alpaca} depict the distribution of the values of \metric\ for all models and variation types in MMLU and Alpaca, respectively.

\end{document}